%% file: main.tex
\documentclass{article}

\PassOptionsToPackage{square,numbers}{natbib}
\usepackage[preprint]{neurips_2021}

\usepackage[utf8]{inputenc} %
\usepackage[T1]{fontenc}    %
\usepackage{hyperref}       %
\usepackage{url}            %
\usepackage{booktabs}       %
\usepackage{amsfonts}       %
\usepackage{nicefrac}       %
\usepackage{microtype}      %
\usepackage{amssymb,amsmath}
\usepackage{subcaption}
\usepackage{wrapfig}
\usepackage{graphicx}
\usepackage{mathtools}
\usepackage{cleveref}
\usepackage{xcolor}
\usepackage{algorithm}
\usepackage{algorithmicx}
\usepackage{algpseudocode}
\usepackage{blindtext}
\usepackage{multirow}
\usepackage[export]{adjustbox}

\newcommand{\card}[1]{|#1|}

\DeclareMathOperator*{\EE}{\mathbb{E}}

\DeclareMathOperator{\entropy}{\mathcal{H}}
\DeclareMathOperator{\targetentropy}{\overline{\mathcal{H}}}
\newcommand{\policy}{\pi}
\newcommand{\policyhi}{\pi^{\rm hi}}
\newcommand{\policylo}{\pi^{\rm lo}}
\newcommand{\policygs}{\pi^{\rm f}}
\newcommand{\policysg}{\pi^{\rm g}}
\newcommand{\ppolicygs}{\pi^{\rm f}_\phi}
\newcommand{\ppolicysg}{\pi^{\rm g}_\psi}

\newcommand{\actions}{\mathcal{A}}
\newcommand{\states}{\mathcal{S}}
\newcommand{\statesppc}{\mathcal{S}^{\rm p}}
\newcommand{\statestask}{\mathcal{S}^+}
\newcommand{\statesgoal}{\mathcal{S}^{\rm g}}
\newcommand{\goals}{\mathcal{G}}

\title{Hierarchical Skills for Efficient Exploration}

\author{%
  \textbf{Jonas Gehring}$^{1,2}$\hspace{1em}
  \textbf{Gabriel Synnaeve}$^1$\hspace{1em}
  \textbf{Andreas Krause}$^2$\hspace{1em}
  \textbf{Nicolas Usunier}$^1$
  \vspace{3pt}
  \\
  $^1$Facebook AI Research\hspace{1em}
  $^2$ETH Z{\"u}rich
  \vspace{3pt}
  \\
  \texttt{jgehring@fb.com}
}

\begin{document}

\maketitle

\begin{abstract}
In reinforcement learning, pre-trained low-level skills have the potential to greatly facilitate exploration.
However, prior knowledge of the downstream task is required to strike the right balance between generality (fine-grained control) and specificity (faster learning) in skill design.
In previous work on continuous control, the sensitivity of methods to this trade-off has not been addressed explicitly, as locomotion provides a suitable prior for navigation tasks, which have been of foremost interest.
In this work, we analyze this trade-off for low-level policy pre-training with a new benchmark suite of  diverse, sparse-reward tasks for bipedal robots.
We alleviate the need for prior knowledge by proposing a hierarchical skill learning framework that acquires skills of varying complexity in an unsupervised manner.
For utilization on downstream tasks, we present a three-layered hierarchical learning algorithm to automatically trade off between general and specific skills as required by the respective task.
In our experiments, we show that our approach performs this trade-off effectively and achieves better results than current state-of-the-art methods for end-to-end hierarchical reinforcement learning and unsupervised skill discovery.
Code and videos are available at \url{https://facebookresearch.github.io/hsd3}.

\end{abstract}

\input{intro}
\input{related-work}

\input{big-picture}

\input{pre-training}

\input{hierarchical-control}

\input{environments}

\input{experiments}
\input{conclusion}

\newpage

\bibliography{main}

\newpage
\appendix
\input{app-envorinments}

\input{app-goal-spaces}
\input{app-pre-training}
\input{app-sac-extension}
\input{app-training-details}
\clearpage
\input{app-hiro-rewards}

\end{document}

%% file: intro.tex
\section{Introduction}
\looseness -1 A promising direction for improving the sample efficiency of reinforcement learning agents in complex environments is to pre-train low-level skills that are then used to structure the exploration in downstream tasks~\citep{konidaris2007building,hausman2018learning,marino2019hierarchical,eysenbach2019diversity,merel2019hierarchicala}. 
This has been studied in particular for the control of (simulated) robots, where there is a natural hierarchical decomposition of the downstream tasks into low-level control of the robot's actuators with a skill policy, and a high-level control signal that specifies a direction or target robot configuration with coarser temporal resolution. 
The large body of work on unsupervised skill or option discovery in hierarchical reinforcement learning (HRL) for continuous control relies, explicitly or implicitly, on prior knowledge that low-level skills should control the center of mass of the robot~\citep{machado2017laplacian,gregor2017variational,eysenbach2019diversity,sharma2020dynamicsaware,campos2020explore}.
This nicely fits a wide range of benchmark tasks that are variants of navigation problems, but the benefit of such hierarchical setups outside this problem class is unclear. 

\looseness -1 The prior knowledge embedded in a pre-trained skill defines a specific trade-off between sample efficiency and generality.
Skills that severely constrain the high-level action space to elicit specific behavior (e.g., translation of the center of mass) are likely to provide the largest gains in sample efficiency, but are unlikely to be useful on a diverse set of downstream tasks.
Conversely, low-level skills that expose many degrees of freedom are more widely applicable but less useful for guiding exploration.
There is, thus, no single universally superior pre-trained skill.
Depending on the downstream task, different skills might also be useful to efficiently explore in different parts of the environment.

In this paper, we aim to acquire skills that are useful for a variety of tasks while still providing strong exploration benefits.
We propose to pre-train a {\em hierarchy of skills} of increasing complexity which can subsequently be composed with a high-level policy.
In the context of simulated robots, each skill consists of controlling a part of the robot configuration over a short time horizon, such as the position of the left foot of a humanoid, or the orientation of its torso.
Skills of increasing complexity are constructed by jointly controlling larger portions of the configuration.
These skills are modelled with a shared policy and pre-trained in an environment without rewards and containing the robot only.
Subsequently, skills are used in downstream tasks within a three-level hierarchical policy: the highest level selects the skill (which specifies a \emph{goal space}), the second level the target configuration within that skill (the \emph{goal}), and the pre-trained skill performs the low-level control to reach the goal. 
Compared to standard approaches involving a single static pre-trained skill \citep{eysenbach2019diversity,sharma2020dynamicsaware}, our approach offers increased flexibility for structuring exploration and offloads the issue of selecting prior knowledge from pre-training to downstream task learning.
As a result, our skills can be acquired once per robot and applied to many different tasks.

We perform an experimental analysis of our hierarchical pre-training on a new set of challenging sparse-reward tasks with simulated bipedal robots.
Our experiments show that each task is most efficiently explored by a distinct set of low-level skills, confirming that even on natural tasks, where locomotion is of primal importance, there is no overall single best pre-trained skill.
We further show that dynamic selection of goal spaces with a three-level hierarchy performs equally or better than a generic skill on all tasks, and can further improve over the best single skills per task.

The main contributions of our work are summarized as follows:
\begin{itemize}
    \item We propose a novel unsupervised pre-training approach that produces a hierarchy of skills based on control of variable feature sets. %
    \item We demonstrate how to automatically select between different skills of varying complexity with a three-level hierarchical policy that selects skills, goals, and native actions.
    \item We introduce a benchmark suite of sparse-reward tasks that allows for consistent and thorough evaluation of motor skills and HRL methods beyond traditional navigation settings.
    \item We study the implications of prior knowledge in skills experimentally and showcase the efficacy of our hierarchical skill framework on the proposed benchmark tasks, achieving superior results compared to existing skill discovery and HRL approaches~\citep{eysenbach2019diversity,nachum2018dataefficient,zhang2020hierarchical}.
\end{itemize}

%% file: related-work.tex
\section{Related Work}
\label{sec:related-work}

The success of macro-operators and abstraction in classic planning systems~\citep{fikes1972learning,sacerdoti1974planning} has inspired a large body of works on hierarchical approaches to reinforcement learning~\citep{dayan1992feudal,sutton1999mdps,bacon2017optioncritic,vezhnevets2017feudal,nachum2018dataefficient,riedmiller2018learning}.
While the decomposition of control across multiple levels of abstraction provides intuitive benefits such as easier individual learning problems and long-term decision making, recent work found that a primary benefit of HRL, in particular for modern, neural-network based learning systems, stems from improved exploration capabilities~\cite{jong2008utility,nachum2019why}.
From a design perspective, HRL allows for separate acquisition of low-level policies (options; skills), which can dramatically accelerate learning on downstream tasks.
A variety of works propose the discovery of such low-level primitives from random walks~\citep{machado2017laplacian,whitney2020dynamicsaware},
mutual information objectives~\cite{gregor2017variational,eysenbach2019diversity,sharma2020dynamicsaware,campos2020explore}, datasets of agent or expert traces~\citep{pertsch2020accelerating,ajay2021opal,li2021planning}, motion capture data~\citep{peng2019mcp,merel2019neural,shankar2020learning}, or from dedicated pre-training tasks~\citep{florensa2017stochastic,marino2019hierarchical}.

In order to live up to their full potential, low-level skills must be useful across a large variety of downstream tasks.
In practice, however, a trade-off between generality (wide applicability) and specificity (benefits for specific tasks) arises.
A large portion of prior work on option discovery and HRL resolved this trade-off, explicitly or implicitly, in favor of specificity.
This can be exemplified by the choice of test environments, which traditionally revolve around navigation in grid-world mazes~\citep{sutton1999mdps,dayan1992feudal,dietterich2000hierarchical}.
In recent work on simulated robotics, navigation problems that are similar in spirit remain the benchmarks of choice~\cite{duan2016benchmarking,nachum2018dataefficient,eysenbach2019diversity,marino2019hierarchical,ajay2021opal}.
In these settings, directed locomotion, i.e., translation of the robot's center of mass, is the main required motor skill, and the resulting algorithms require a corresponding prior.
This prior is made explicit with skills partitioning the state space according the agent's position~\citep{eysenbach2019diversity,sharma2020dynamicsaware}, or is implicitly realized by high contribution of position features to reward signals~\citep{nachum2018dataefficient,nachum2019nearoptimal,ajay2021opal} (\Cref{sec:hiro-reward-normalization}).
Similarly, works that target non-navigation environments acquire skills with tailored pre-training tasks~\citep{qureshi2020composing} or in-domain motion capture data~\citep{peng2019mcp,merel2019neural,merel2020catch}.
In contrast, our work is concerned with learning low-level skills without extra supervision from traces or pre-training task design, and which do not prescribe a fixed trade-off towards a particular type of behavior.

%% file: big-picture.tex
\section{Hierarchical Skill Learning}
\label{sec:big-picture}

\subsection{Overview}

\begin{figure}
  \centering
  \includegraphics[width=\textwidth,trim={1em 1.5em 1em 1em},clip]{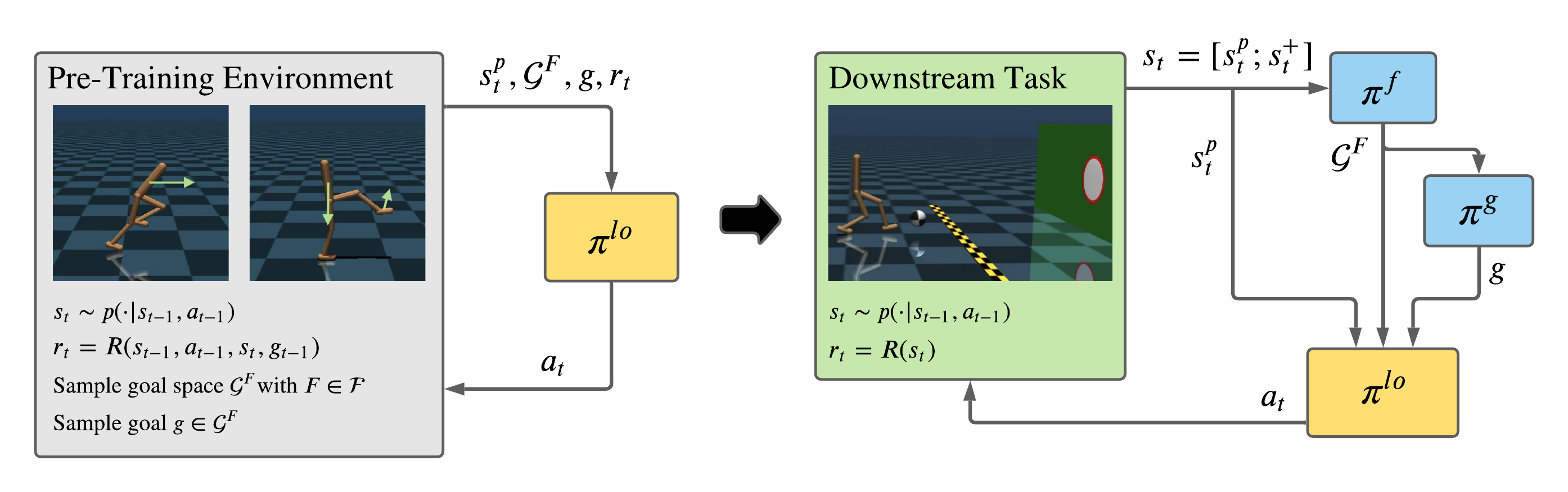}
  \caption{
  \looseness -1 Illustration of our proposed hierarchical skill learning framework.
  \textbf{Left} Low-level policies are learned in an empty pre-training environment, with the objective to reach random configurations (goal $g$) of a sampled skill (goal space $\mathcal{G}^F$ defined over a feature set $F$).
  Examples of goal space features are translation along the X-axis or the position of a limb.
  \textbf{Right} Learning on downstream tasks with a three-level hierarchical policy to select a goal space, a goal and finally a native action $a_t$ with the pre-trained low-level policy.
  The low-level policy acts on proprioceptive states $s^p$, while high-level policies $\policygs$ and $\policysg$ leverage extra task-specific information via $s^+$.}
  \label{fig:hsd3-system}
\end{figure}

\looseness -1 In this work, we propose an approach where we first acquire low-level policies that are able to carry out a useful set of skills in an unsupervised manner, i.e., without reward signals from the main tasks of interest. We subsequently employ these skills in a hierarchical reinforcement learning setting (\Cref{fig:hsd3-system}).
The pre-training environment consists solely of the robot.
In downstream tasks, additional objects may be present besides the robot; these may be fixed (e.g., obstacles) or only indirectly controllable (such as a ball).

\looseness -1 Crucially, the usefulness of a skill policy stems from providing additional structure for effective exploration.
In order to provide benefits across a wide range of tasks, skills need to support both fast, directed exploration (e.g., locomotion) as well as precise movements (e.g., lift the left foot while bending down).
We propose to fulfill these requirements with short-horizon, goal-directed low-level policies that are trained to achieve target configurations of robot-level state features such as the position or orientation of its torso or relative limb positions.
We denote this feature space with $\statesgoal$.
This allows the definition of a hierarchy of skills by controlling single features and their combinations, resulting in varying amounts of control exposed to high-level policies.
Each skill is trained to reach goals in a goal space $\mathcal{G}^F$ defined over a set of features $F$ of $\statesgoal$, yielding policies of the form
\begin{equation*}
    \left\{ \policylo_F: \statesppc \times \mathcal{G}^F \to A \right\}_{F \in \mathcal{F}}
\end{equation*}
with proprioceptive observations $\statesppc$ (\Cref{fig:hsd3-system}, left).
Goal-directed skill policies are trained without task-specific rewards, relying solely on state features related to the specific robot and the prescribed hierarchy of goal spaces.

\looseness -1 On downstream tasks, high-level policies operate with a combinatorial action space as the skill policy $\policylo$ is conditioned on both a feature set $F$ and a concrete goal $g \in \mathcal{G}^F$.
State spaces are enriched with task-specific features $\statestask$, s.t. $\states = \statesppc \cup \statestask$, that contain information regarding additional objects in the downstream task. This enriched state space is only available to the high-level policy.
We model the high-level policy in a hierarchical fashion with two policies %
\begin{equation*}
    \policygs: S \to \mathcal{F},\quad \policysg: S \times \mathcal{F} \to \mathcal{G}^F,
\end{equation*}
that prescribe a goal space and a goal, respectively (\Cref{fig:hsd3-system}, right).
With a task- and state-dependent policy $\policygs$, it is possible to not only select the required set of skills for a given environment, but also to switch between them within an episode. In this three-level hierarchy, the higher-level policy $\policygs$ explores the hierarchy of goal spaces and dynamically trades off between generality and specificity.
Temporal abstraction is obtained by selecting new high-level actions at regular intervals but with a reduced frequency compared to low-level actions.
We term the resulting learning algorithm \textit{HSD-3}, emphasizing the three different levels of control (goal spaces, goals and native actions) obtained after our hierarchical skill discovery phase.

%% file: pre-training.tex
\subsection{Unsupervised Pre-Training}
\label{sec:unsup-pre-training}

During pre-training, skill policies act in an MDP $\mathcal{M}$ with proprioceptive states $s^p$, native actions $a$, transition probabilities and initial states $s^p_0 \in \states^p_0$ drawn from a distribution $P_0$~\citep[for example]{sutton2018reinforcement}, but without an extrinsic reward function.
The skill policies are trained to reach goals in a goal space defined over $S^g$ consisting of robot-level sensor readings, which may include non-proprioceptive information.

Critically, each skill $\policylo_F$ aims to achieve goals defined over a variable-sized subset of features $F$ of $S^g$.
To avoid learning a combinatorial number of independent policies, we follow the principle of universal value function approximators~\citep{schaul2015universal} and share the parameters of each skill. We augment the policy's input with goal and goal space information, and learn a single set of parameters $\theta$, leading to $\pi^{lo}_\theta: \statesppc \times \mathcal{F} \times \mathcal{G} \to \actions$, where $\mathcal{F} \coloneqq \{F: F \subseteq \statesgoal\}$ and $\goals \coloneqq \{{\cal G}^F: F \subseteq \statesgoal\}$. For ease of modeling, $F$ is provided to $\pi^{lo}_\theta$ as a bag-of-words input where a coordinate is set to 1 if its respective feature is included in $F$ and to 0 otherwise.
Also, the low-level policy receives goals relative to the current values of the $s^{\rm g}$. Thus, at each time step, the input goal to the policy is updated to reflect the progress towards the goal.
As a learning signal, we provide a distance-based reward as $R(s^g,a,s^{g'},F,g) \coloneqq ||\omega^F(s^g) - g||_2 - ||\omega^F(s^{g'}) - g||_2$, where
$\omega^F: \statesgoal \to {\cal G}^F$ is a fixed transformation that selects the subset $F$ of goal space features and applies a suitable normalization.
The selection of features for $\statesgoal$ and transformations $\omega^F$ represents the prior knowledge that we utilize for unsupervised pre-training.
We discuss examples for bipedal robots in our experiments and in~\Cref{sec:robots-and-goal-spaces}.

During training, features $F$ and goals $g$ are sampled anew for each episode in the environment.
Episodes consist of several dozen steps only, reflecting the fact that we are interested in short-horizon skills.
In-between episodes, we reset the simulation to states drawn from $\mathcal{S}^p_0$ only sporadically to encourage the resulting policies to be applicable in a wide range of states, facilitating their sequencing under variable temporal abstractions.
Likewise, rewards are propagated across episodes as long as the simulation is not reset.
The low-level policy parameters $\theta$ can be optimized with any reinforcement learning algorithm; here, we opt for Soft Actor-Critic (SAC), a state-of-the-art method tailored to continuous action spaces~\citep{haarnoja2018soft}.
Pseudo-code for the pre-training algorithm and further implementation details are provided in~\Cref{sec:unsup-pre-training-details}.

%% file: hierarchical-control.tex
\subsection{Hierarchical Control}
\label{sec:hierarchical-control}

\looseness -1 After a low-level policy $\policylo$ has been obtained via unsupervised pre-training, we employ it to construct a full hierarchical policy
$\policy(a|s) = \policygs(F|s) \policysg(g|s,F) \policylo(a|s^p,F,g)$.
The hierarchical policy is applied to downstream tasks, where we are supplied with a reward signal and state observations $\states = \statesppc \cup \statestask$.
While different frequencies may be considered for acting with $\policygs$ and $\policysg$, in this work both high-level actions are selected in lock-step.
For jointly learning $\policygs$ and $\policysg$, we extend the SAC algorithm to incorporate the factored action space $\mathcal{F} \times \goals$ . In particular, it is necessary to extend SAC to deal with a joint discrete and continuous action space, which we describe below (see \Cref{sec:sac-extension} for further details and pseudo-code for high-level policy training).

We formulate our extension with a shared critic $Q(s,F,g)$, incorporating both factors of the action space.
$Q(s,F,g)$ is trained to minimize the soft Bellman residual~\citep[Eq. 5]{haarnoja2018soft}, with the soft value function $V(s)$ including entropy terms for both policies.
Since $\policygs$ outputs a distribution over discrete actions (to select from available goal spaces), we compute its entropy in closed form instead of using the log-likelihood of the sampled action as for $\policysg$.
We employ separate temperatures $\alpha$ and $\beta^F$ for the goal space policy as well as for each corresponding goal policy; $\beta^F$ are normalized to account for varying action dimensions across goal spaces.
\begin{equation*}
V(s) = \sum_{F \in \mathcal{F}} \policygs(F | s)  \EE_{g \sim \policysg} \left[ Q(s,F,g) - \frac{\beta^F}{|F|} \log \policysg(g | s,F) \right]  + \alpha \left( \entropy(\policy^f(\cdot|s)) - \log |\mathcal{F}| \right).
\end{equation*}
Compared to the usual SAC objective for continuous actions, we subtract $\log\card{\cal F}$ from the entropy to ensure a negative contribution to the reward. This matches the sign of the entropy penalty from $\policysg$, which is negative in standard hyper-parameter settings\footnote{The entropy for continuous actions sampled from a Gaussian distribution can be negative since it is based on probability densities.}~\citep{haarnoja2018soft}. Otherwise, the entropy terms for discrete and continuous actions may cancel each other in Q-function targets computed from $V(s)$.

The policy loss~\citep[Eq. 7]{haarnoja2018soft} is likewise modified to include both entropy terms.
Parameterizing the policies with $\phi$ and $\psi$ and the Q-function with $\rho$, we obtain
\begin{equation*}
J_{\policy^{f,g}}(\phi,\psi) = \EE_{s \sim B} \left[ \sum_{F \in \mathcal{F}} \ppolicygs(F | s) \EE_{g \sim \ppolicysg} \left[ \frac{\beta^F}{|F|} \log \ppolicysg(g | F, s) - Q_\rho(s,F,g) \right] - \alpha \entropy(\ppolicygs(\cdot|s)\right],
\end{equation*}
where the replay buffer is denoted with $B$.
The scalar-valued entropy temperatures are updated automatically during learning to maintain target entropies $\targetentropy^f$ and $\targetentropy^g$ for both $\policygs$ and $\policysg$.
The following losses are minimized at each training step:
\begin{align*}
J(\alpha) &= \EE_{s \sim B} \left[ \alpha \left( \entropy(\policygs(\cdot|s)) - \targetentropy^f \right)  \right] \\
J(\beta) &= \EE_{s \sim B} \left[ -\sum_{F \in \mathcal{F}} \beta^F \policygs(F | s) \EE_{g \sim \policysg} \left[ \frac{1}{|F|} \log \policysg(g | s,F) + \targetentropy^g \right] \right].
\end{align*}

We implement temporal abstraction by selecting high-level actions with a reduced frequency.
This incurs a reduction in available training data since, when taking a high-level action every $c$ steps in an $N$-step episode, we obtain only $N/c$ high-level transitions.
To leverage all available transitions gathered during training, we adopt the step-conditioned critic proposed by \citet{whitney2020dynamicsaware} in our SAC formulation.
The Q-function receives an additional input $0 \leq i \leq c$ marking the number of steps from the last high-level action, and is trained to minimize a modified soft Bellman residual:
\begin{align*}
J_Q(\rho) = \EE_{\substack{F_t,g_t,i,\\s_{t,\dots,t+c-i},\\a_{t,\dots,t+c-i} \sim B}} \left[ \frac{1}{2} \left( Q_\rho(s_t,F_t,g_t,i) - \left( \sum_{j=0}^{c-i-1} \left( \gamma^j r(s_{t+j},a_{t+j}) \right) + \gamma^{c-i} V(s_{t+c-i}) \right) \right)^2 \right]
\end{align*}
As in~\citep{whitney2020dynamicsaware}, $Q(s,F,g,0)$ is used when computing $V(s)$ and the policy loss $J_{\policy^{f,g}}(\phi,\psi)$.

The factorized high-level action space of HSD-3 has previously been studied as parameterized action spaces~\citep{masson2016reinforcement,wei2018hierarchical}, albeit with a small number of discrete actions and not in the context of hierarchical RL.
A possible application of Soft Actor-Critic is described in~\citet{delalleau2019discrete}; our approach differs in that we (a) compute the soft value function as a weighted sum over all discrete actions, and (b) opt for two separately parameterized actors and a shared critic.
These design choices proved to be more effective in initial experiments.

%% file: environments.tex
\section{Benchmark Environments}
\label{sec:environments}

\begin{figure}
  \centering
  \begin{subfigure}[b]{0.16\linewidth}
    \centering
    \includegraphics[width=\linewidth]{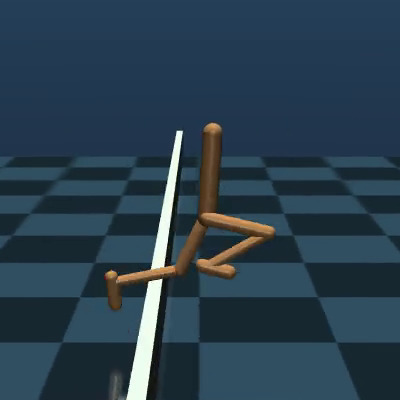}
    \caption{Hurdles}
    \label{fig:task-hurdles}
  \end{subfigure}
  \begin{subfigure}[b]{0.16\linewidth}
    \centering
    \includegraphics[width=\linewidth]{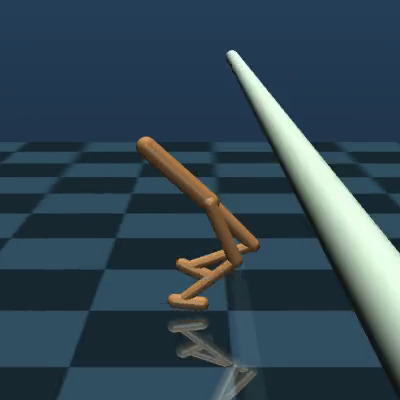}
    \caption{Limbo}
    \label{fig:task-limbo}
  \end{subfigure}
  \begin{subfigure}[b]{0.16\linewidth}
    \centering
    \includegraphics[width=\linewidth]{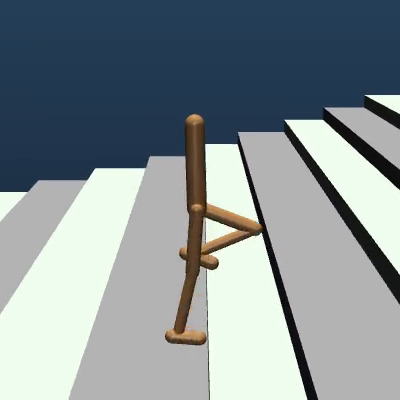}
    \caption{Stairs}
    \label{fig:task-stairs}
  \end{subfigure}
  \begin{subfigure}[b]{0.16\linewidth}
    \centering
    \includegraphics[width=\linewidth]{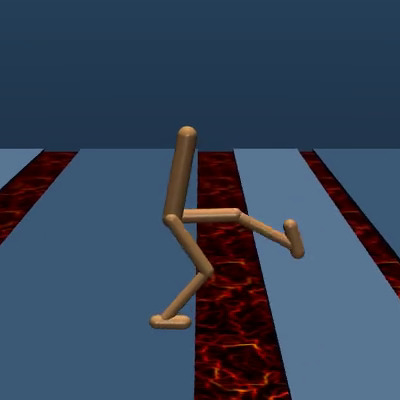}
    \caption{Gaps}
    \label{fig:task-gaps}
  \end{subfigure}
  \begin{subfigure}[b]{0.16\linewidth}
    \centering
    \includegraphics[width=\linewidth]{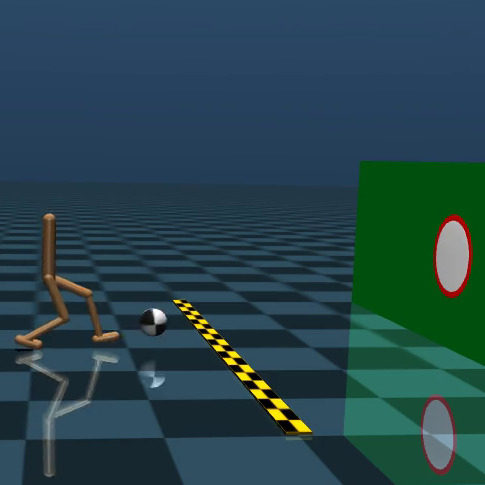}
    \caption{GoalWall}
    \label{fig:task-goalwall}
  \end{subfigure}
  \begin{subfigure}[b]{0.16\linewidth}
    \centering
    \includegraphics[width=\linewidth]{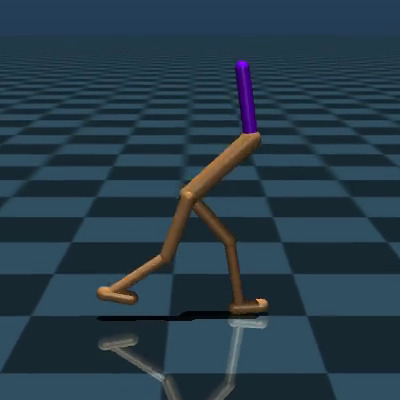}
    \caption{PoleBalance}
    \label{fig:task-polebalance}
  \end{subfigure}
  \caption{Benchmark environments for evaluating motor skills for bipedal robots, pictured with the Walker robot. Hurdle and Limbo bar heights and spacing, as well as stair lengths, are sampled randomly from fixed distributions. We also include a combination of Hurdles and Limbo (HurdesLimbo), in which both obstacle types alternate.}
  \label{fig:tasks}
\end{figure}

\looseness -1 We propose a benchmark suite for comprehensive evaluation of pre-trained motor skills, tailored to bipedal robots (\Cref{fig:tasks}) and implemented for the MuJoCo physics simulator~\citep{todorov2012mujoco,tassa2020dm}.
We selected tasks that require diverse abilities such as jumping (Hurdles), torso control (Limbo), fine-grained foot control (Stairs, GoalWall) and body balance (PoleBalance).
Besides PoleBalance, all tasks also require locomotion.
The tasks are designed to have a sparse reward that is provided once an obstacle or step has been passed or a goal has been shot, emphasizing the need for good exploration.
As an exception, PoleBalance provides a constant reward during the entire episode, which ends when the pole placed on the robot falls over.
PoleBalance requires precise motor control, and is, in combination with its reward structure, chiefly outside of the realm of benchmarks that have been traditionally used to evaluate HRL techniques. 
Obstacle position and step lengths are subject to random perturbations, sampled anew for each episode.
We tailor several environment parameters to the concrete robot being controlled, e.g., the positions of Limbo bars.
For the effectively two-dimensional Walker robot, we limit the movement of objects (ball, pole) to translation along X and Z and rotation around the Y-axis.
In all environments, we label simulation states that would cause the robot to fall over as invalid; these will terminate the episode with a reward of -1.
\Cref{sec:environment-details} contains full specifications for all tasks.

Variants of some of our benchmark tasks have been proposed in prior works: Hurdles~\citep{heess2017emergence,li2019subpolicy,eysenbach2019diversity,qureshi2020composing}, Stairs~\citep{merel2017learning,ghosh2018divideandconquer}, Gaps~\citep{heess2017emergence,tassa2020dm}, and PoleBalance is inspired by the classical CartPole problem~\citep{barto1983neuronlike}.
Our tasks are different from earlier published counterparts in parameterization and in the sparsity of rewards they provide.
All environments are provided via a standard Gym interface~\citep{brockman2016openai} with accompanying open-source code, enabling easy use and re-use.

%% file: experiments.tex
\section{Experimental Results}
\label{sec:experiments}

For our experiments, we first pre-train skill policies as described in~\Cref{sec:unsup-pre-training} in an empty environment.
The feature set to describe target configurations consists of the robot's translation along the X-axis, its position on the Z-axis, torso rotation around the Y-axis and positions of each foot, relative to the body~(\Cref{sec:robots-and-goal-spaces}).
With these five features, we obtain a set of $2^5-1=31$ skills with a shared set of policy parameters.
We then train separate high-level policies for each benchmark environment, directing the same set of pre-trained skills.
High-level actions are taken every 5 environment steps; in the PoleBalance environment, which requires short reaction times, all three policies operate at the same frequency.
All neural network policies employ skip connections as proposed for SAC by~\citet{sinha2020d2rl}.
We show results over 9 seeds for each run; in experiments involving pre-training, we combine 3 seeds for pre-training with 3 seeds for high-level policy training.
Pre-training takes approximately 3 days on 2 GPUs (V100) and fine-tuning (downstream task training) takes 2 days to reach 5M samples on 1 GPU. 
Further details regarding the training setup, hyper-parameters for skill and high-level policy training, as well as for baselines, can be found in~\Cref{sec:training-details}.

\subsection{Trade-offs for Low-Level Skills}
\label{sec:exp-gs-tradeoff}

\begin{figure}[b]
  \centering
  \includegraphics[width=\textwidth]{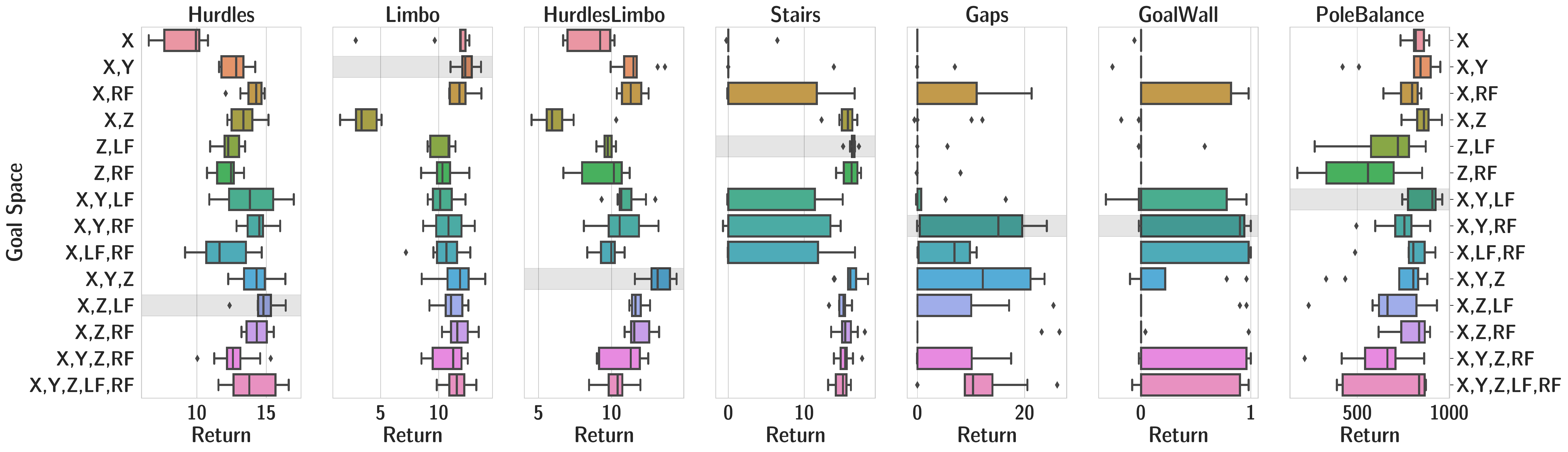}
  \caption{Returns achieved after 5M samples on the benchmark tasks with the Walker robot with fixed low-level policy goal spaces (quartiles). Each row corresponds to a set of features for the respective goal space. Best skills (marked) differ significantly across tasks.}
  \label{fig:walker-single}
\end{figure}

In a first experiment with the Walker robot, we aim to investigate the trade-off between specificity and generality of low-level skills.
For each environment, we train high-level policies directing a {\em fixed} skill within its respective goal space, which for example consists of torso-related features only, or of feet positions combined with forward translation.
At the end of training, we measure average returns obtained for all 31 skills in a deterministic evaluation setting, and plot quartiles over different seeds for the best skills in~\Cref{fig:walker-single}.
The performance for different fixed skills varies significantly across environments, and {\em no single skill is able to obtain the best result (shaded) across all environments}.
While controlling X, Y and Z features only produces the good hierarchical policies for most tasks, poor results are obtained in Gaps and GoalWall.
Controlling feet positions (LF and RF) is crucial for GoalWall, although close-to-optimal performance is only achieved in few runs across the respective skills.
The most general skill, high-level control of {\em all} considered features (bottom row), also allows for learning progress on all tasks; however, it is generally outperformed by other, more specific skills.

\subsection{Skill Selection with HSD-3}
\label{sec:walker-results}

We compare HSD-3 against a number of base- and toplines to examine the effectiveness of our proposed framework for skill learning and hierarchical control.
We include state-of-the-art  non-hierarchical RL methods (SAC~\citep{haarnoja2018soft}), end-to-end hierarchical RL algorithms (HIRO~\citep{nachum2018dataefficient}, HIDIO~\cite{zhang2020hierarchical}),
and evaluate DIAYN as an alternative unsupervised pre-training algorithm~\citep{eysenbach2019diversity}.
For HIRO, we use SAC for learning both high- and low-level policies (HIRO-SAC).
We facilitate the comparison to DIAYN with a continuous-valued skill variable obtained from an embedding~\citep{achiam2018variational} (DIAYN-C).
For both HIRO and DIAYN-C, we provide the same goal space as in our pre-training stage, consisting of the full feature set, to ensure a fair comparison.
We further compare to the Switching Ensemble proposed in~\cite{nachum2019why}, which does not learn a high-level policy but has been shown to improve exploration.
Finally, we examine how our pre-trained skills perform if we use a single goal space only (SD).
Without prior knowledge about the task at hand, this corresponds to the full goal space $F = \{0,1,...,\dim(\statesgoal)\}$, settling for maximum generality.
As a topline, we select the \emph{best per-task} goal spaces from~\Cref{sec:exp-gs-tradeoff}, denoted with SD*, which required exhaustive training and evaluation with all available goal spaces.

\begin{figure}
  \centering
  \includegraphics[width=\textwidth]{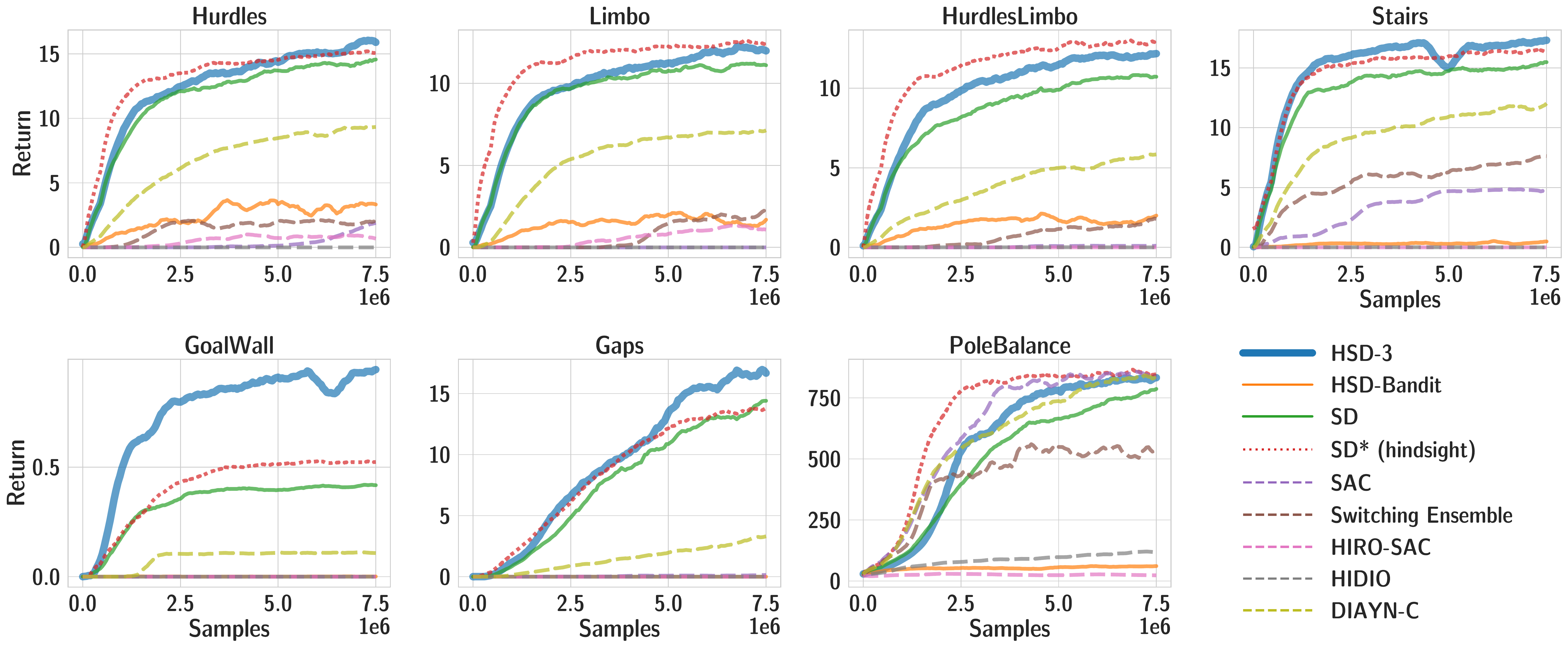}
  \caption{Learning curves on benchmark environments with the Walker robot.
  For clarity, we plot mean performance over 9 seeds, averaged over 0.5M samples.
  Full learning curves including error bands are provided in~\Cref{sec:learning-curves-walker}.}
  \label{fig:walker-perf}
\end{figure}

\setlength{\tabcolsep}{5pt}
\begin{table}[b]
  \centering
  \small
  \resizebox{\textwidth}{!}{
  \begin{tabular}{lrrrrrrr}
    \toprule
    Method & Hurdles & Limbo & HurdlesLimbo & Stairs & Gaps & GoalWall & PoleBalance \\
    \midrule
    SAC & 2.2$\pm$6.2 & -0.1$\pm$0.2 & -0.0$\pm$0.4 & 5.0$\pm$4.8 & 0.1$\pm$0.5 & -0.2$\pm$0.3 & \textbf{866.8}$\pm$104.5 \\
    Switching Ensemble & 0.7$\pm$3.0 & 1.9$\pm$4.3 & 1.6$\pm$3.6 & 7.4$\pm$3.8 & -0.2$\pm$0.3 & -0.2$\pm$0.3 & 569.4$\pm$230.1 \\
    HIRO-SAC & 0.4$\pm$1.6 & 0.9$\pm$2.2 & -0.0$\pm$0.1 & -0.0$\pm$0.0 & -0.1$\pm$0.2 & -0.0$\pm$0.0 & 23.8$\pm$12.4 \\
    HIDIO & -0.1$\pm$0.1 & -0.1$\pm$0.1 & -0.2$\pm$0.1 & -0.2$\pm$0.3 & -0.2$\pm$0.3 & -0.3$\pm$0.3 & 117.6$\pm$33.8 \\
    DIAYN-C & 9.5$\pm$2.5 & 7.5$\pm$1.3 & 5.7$\pm$1.7 & 12.5$\pm$2.9 & 3.4$\pm$4.1 & 0.1$\pm$0.3 & \textbf{839.9}$\pm$58.4 \\
    \midrule
    SD & \textbf{15.0}$\pm$1.4 & 11.0$\pm$1.0 & 10.8$\pm$1.4 & 15.2$\pm$0.9 & \textbf{14.5}$\pm$9.3 & 0.4$\pm$0.5 & 789.5$\pm$79.1 \\
    HSD-Bandit & 2.6$\pm$1.3 & 2.4$\pm$1.7 & 2.0$\pm$1.5 & 0.5$\pm$0.8 & -0.1$\pm$0.2 & -0.1$\pm$0.2 & 61.0$\pm$15.8 \\
    HSD-3 & \textbf{15.3}$\pm$2.0 & \textbf{11.7}$\pm$0.9 & \textbf{11.8}$\pm$1.3 & \textbf{17.2}$\pm$0.7 & \textbf{15.1}$\pm$8.9 & \textbf{0.9}$\pm$0.1 & \textbf{876.0}$\pm$36.9 \\
    \midrule
    SD* & \textbf{15.0}$\pm$1.0 & \textbf{12.2}$\pm$1.0 & \textbf{12.7}$\pm$1.3 & 16.2$\pm$1.3 & \textbf{14.0}$\pm$11.3 & 0.5$\pm$0.5 & \textbf{868.2}$\pm$91.1 \\\specialrule{\heavyrulewidth}{\aboverulesep}{2\belowrulesep}
    \end{tabular}
  }
  \caption{Final performance after 7.5M samples across benchmark tasks with the Walker robot. Mean and standard deviation reported for average returns over 9 seeds.}
  \label{tab:walker-perf}
\end{table}

In~\Cref{fig:walker-perf}, we plot learning curves over 5M training interactions for HSD-3 as well as base- and toplines, and list final performances in~\Cref{tab:walker-perf}.
On all tasks, HSD-3 outperforms or matches (Limbo) the learning speed of a conventional hierarchical policy with a skill defined over the full goal space (SD).
On GoalWall and Stairs, the {\em three-level policy significantly exceeds the mean performance of single goal space base- and toplines}.
For Gaps, the variances of the best results in~\Cref{tab:walker-perf} (HSD-3, SD, SD*) are high, hinting at especially challenging exploration problems.
A closer look at per-seed performance (\Cref{sec:learning-curves-walker}) reveals learning failure for 2 out of 9 runs for these methods.
For GoalWall, 4 and 5 runs for SD* and SD, respectively, achieve zero return.
Per-task goal space selection with a bandit algorithm (HSD-Bandit) fails on most environments, and achieves low returns on Hurdles, Limbo and HurdlesLimbo.

\begin{figure}[t]
  \centering
  \includegraphics[width=\textwidth]{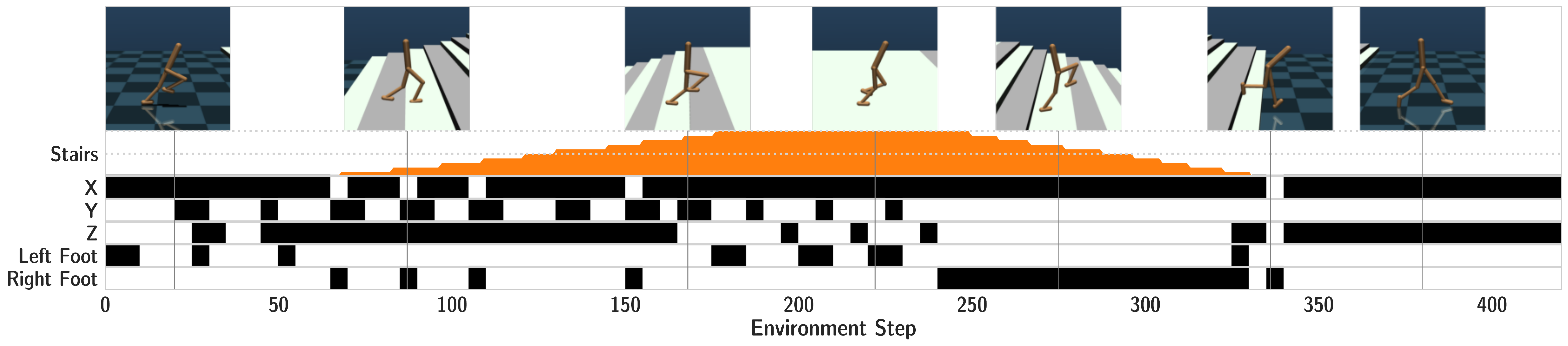}
  \caption{Skills selected by HSD-3 during an episode on the Stairs task.
  Different skills are utilized for different sections of the environment: walking upstairs is achieved by controlling the torso position (X,Z) in a regular pattern, with goals for the rotation (Y) and the right foot being set occasionally.
  On the top section, a different pattern is selected for quickly running forwards.
  When going downstairs, the right foot of the robot is controlled explicitly to maintain balance.
  }
  \label{fig:walker-trace-stairs}
\end{figure}

These results demonstrate that, without additional task-specific knowledge, our hierarchical skill framework is able to {\em automatically and efficiently trade off between skills of varying complexity at training time}.
In~\Cref{fig:walker-trace-stairs}, we visualize the goal spaces that are selected across the course of an episode in the Stairs environment.
During different stages of the environment (walking upstairs, downstairs, and on a planar surface), HSD-3 selects appropriate skills of varying complexity\footnote{Videos are available at \url{https://facebookresearch.github.io/hsd3}.}.

The performance of the remaining baselines underscores the challenges posed by our benchmark tasks.
Standard SAC works well in PoleBalance, which has a dense reward structure compared to the remaining environments.
On most other tasks, learning progress is slow or non-existent, e.g., on Hurdles, only 2 out of 9 runs achieve a positive return at the end of training.
DIAYN-C exhibits subpar performance across compared our hierarchical pre-training scheme, with the exception of PoleBalance where learning is fast and returns are higher than for SD.
The end-to-end HRL methods HIRO-SAC and HIDIO are unable to make meaningful progress on any task, which highlights the utility of skill pre-training in the absence of task-specific priors.
The Switching Ensemble is the best baseline without pre-training (except for the PoleBalance task), but a clear majority of individual runs do not obtain positive returns (\Cref{sec:learning-curves-walker}).
On the Stairs task, both SAC and the Switching Ensemble manage to climb the flight of stairs in several runs, but fail to discover the second, downwards flight at the end of the platform.

We refer to the supplementary material for further ablation and analysis with the Walker robot.
In \Cref{sec:ablation-pre-training}, we demonstrate the efficacy of our proposed hierarchical pre-training method, compared to pre-training a single skill policy on the full goal space exclusively.
In ~\Cref{sec:exploration-behavior}, we analyze the exploration the behavior of various methods from~\Cref{tab:walker-perf}.
We find that, in general, hierarchical methods visit more states compared to SAC.
HIRO-SAC and DIAYN-C visit a higher number of states in the Hurdles, Limbo, HurdlesLimbo and Stairs environments but fail to leverage this experience for achieving higher returns.

\subsection{Evaluation on a Humanoid Robot}
\label{sec:exp-humanoid}

To evaluate the scalability of HSD-3 to more complex robots, we perform an additional set of experiments with the challenging 21-joint Humanoid robot from~\citet{tassa2020dm}.
For skill pre-training, we select a goal space approximately matching the Walker setup, consisting of the translation along the X-axis of simulation, Z position of the torso, its rotation around the Y axis and three-dimensional feet positions relative to the robot's hip location.

\begin{figure}[t]
  \centering
  \includegraphics[width=\textwidth]{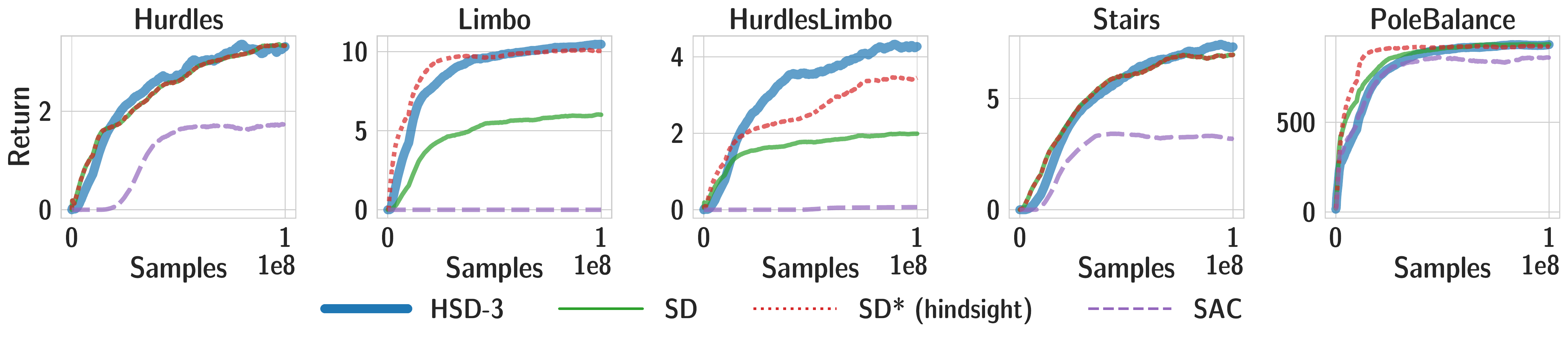}
  \caption{Results on four of the benchmark tasks with a 21-joint Humanoid robot.
  We show mean performance over 9 seeds, clipped to 0 and averaged over 2.5M samples. \Cref{sec:learning-curves-humanoid} contains learning curves for all seeds and error bands.}
  \label{fig:humanoid-results}
\end{figure}

In~\Cref{fig:humanoid-results}, we plot average learning progress across 100M environment steps, comparing HSD-3 to SD (the full goal space), SD* (the best goal space), and a conventional, non-hierarchical SAC setup.
The results align with the observations for the Walker robot in~\Cref{sec:walker-results} in that HSD-3  matches or surpasses the performance of the full goal space, and can further outperform the best goal space selected in hindsight.
Both HSD-3 and SD make good learning progress on three out of five tasks used for this evaluation (Hurdles, Stairs, PoleBalance); on Hurdles and Stairs, SD is in fact the best single goal space.
On HurdlesLimbo, which requires both jumping and crouching, HSD-3 outperforms both SD and SD*.
On the Limbo task, HSD-3 is able to learn faster than SD and achieves a higher return by utilizing the most effective single skills on this task, which control X translation and leaning forward or backward.
This behavior was not discovered by the SD baseline, which exercises control over all goal space features.
In this preliminary study, none of the hierarchical policies was able to make significant progress on the GoalWall and Gaps tasks, which were already shown to be difficult with the Walker robot in~\Cref{sec:walker-results}. 

%% file: conclusion.tex
\section{Conclusion}
\label{sec:conclusion}

\looseness -1 Our work is among the first to highlight the specific trade-offs when embedding prior knowledge in pre-trained skills, and to demonstrate their consequences experimentally in simulated robotic control settings.
We describe a hierarchical skill learning framework that, compared to existing approaches, allows high-level policies to perform the trade-off between directed exploration and fine-grained control automatically during training on downstream tasks.
Our experiments with a bipedal Walker robot demonstrate HSD-3's efficacy on a variety of sparse-reward tasks in which previous approaches struggle.
We further apply our framework to a challenging Humanoid robot, where it learns effective policies on the majority of tasks.

With this work, we provide a new set of benchmark tasks that require diverse motor skills and pose interesting challenges for exploration.
We release the tasks in a dedicated distribution, with the aim of spurring further research on motor skills in continuous control settings, and to ultimately broaden the utility of hierarchical approaches to control to tasks beyond navigation.

\paragraph{Limitations:}  
Our investigations with the Humanoid robot have been performed in limited goal spaces and hence with significant priors.
We believe that further work is required to successfully acquire a large, exhaustive set of low-level skills in unsupervised environments that work on more and more complex morphologies.

\paragraph{Acknowledgements:}
We thank Alessandro Lazaric for insightful discussions, and Franziska Meier, Ludovic Denoyer, and Kevin Lu for helpful feedback on early versions of this paper.

%% file: app-envorinments.tex
\section{Environment Details}
\label{sec:environment-details}

For all benchmark tasks, we place the robot at a designated starting position and configuration, and perturb its joint positions with noise sampled uniformly from $[-0.1,0.1]$ and its joint velocities with noise sampled from $0.1 \cdot \mathcal{N}(0,1)$.
These perturbations are also applied in standard MuJoCo benchmark tasks~\cite{brockman2016openai}.
Each environment provides three different observations: proprioceptive robot states $\statesppc$, extra task-specific observations $\statestask$, and measurements for goal states $\statesgoal$.
Goal state features are solely used for convenience when providing relative goal inputs to low-level policies (\Cref{al:high-level}), and can also be derived from proprioceptive observations and the robot's absolute position.
Below, we list detailed environment configurations and reward functions, as well as robot-specific modifications (if applicable).
In all cases, we define invalid states on a per-robot basis (\Cref{sec:robots-and-goal-spaces}) that lead to a premature end of the episode with a reward of $-1$.
Unless otherwise noted, episodes consist of 1000 steps of agent interaction.

\textbf{Hurdles} Hurdles take the form of simple boxes, placed in intervals $\sim \mathcal{U}(3,6)$ meters, and with heights $\sim \mathcal{U}(0.1,0.3)$ meters.
Task-specific observations are the distance to the next hurdle and its height.
For every hurdle that the robot's torso passes, the agent receives a reward of 1.

\textbf{Limbo} Cylindrical bars with a diameter of 0.2m are placed in intervals $\sim \mathcal{U}(3,6)$ meters.
Their heights are draw from $\mathcal{U}(1.2,1.5)$ for the Walker robot, and from $\mathcal{U}(0.9,1.2)$ for the Humanoid.
The agent observes the distance to the next limbo bar, as well as its height above the ground.
A reward of 1 is obtained if the agent's torso position moves past a bar.

\textbf{HurdlesLimbo} This is a combination of the Hurdles and Limbo environments, where hurdles and limbo bars alternate, starting with a hurdle.
The reward is defined as above, and the observation consists of the type of obstacle (0 for a hurdle, 1 for a limbo bar) as well as its relative distance and height.

\textbf{Stairs} This task consists of 10 stairs (upwards), a platform, and another 10 stairs (downwards).
Stairs have a height of 0.2m, and their length is sampled uniformly within [0.5m,1.0m].
The agent observes the distance of the robot's torso to the next two steps, and a reward of 1 is provided whenever it moves past a step.

\textbf{Gaps} We sample gaps from $\mathcal{U}(0.2,0.7)$ meters and platforms from $\mathcal{U}(0.5,2.5)$ meters ($\mathcal{U}(1.0,2.5$ meters for the Humanoid).
Gaps and platforms are placed in alternation, starting at a distance of 4 meters from the robot's initial position.
Gaps are placed 5cm below the platforms, and the episode ends with a reward of $-1$ if the robot touches a gap.
The agent observes the relative distances to the start of the next gap and next platform.
A reward of 1 is achieved if the robot's torso moves past the start of a platform for the first time.

\textbf{GoalWall} A ball is placed 2.5 meters in front of the robot (1m for the Humanoid).
A wall is placed 4 meters from the ball, and contains a circular target area with an 0.8m diameter, placed 1m above ground.
For the effectively two-dimensional Walker robot, ball movements are restricted to rotation around the Y-axis and translation along X and Z.
At the start of the episode, the ball's position is perturbed with additive noise drawn from $\mathcal{N}(0,0.01)$, and we add noise from $\mathcal{N}(0,0.1)$ to its rotation.
The agent observes the position of the ball (X and Y are relative to the robot's torso, Z is absolute) and its velocity.
Episodes end after 250 steps or if the ball's X position reaches the position of the wall, with a reward of 1 if the ball is inside the target area and 0 otherwise.

\textbf{PoleBalance} A cylindrical pole with a mass of 0.5kg is attached to the robot's topmost body part.
The pole has a length of 0.5m, and the position and velocity of the rotational joint that connects it to the robot is perturbed with additive noise from $\mathcal{N}(0,0.01)$.
For the Walker robot, the pole can rotate around the Y-axis only; for the Humanoid robot, rotations around all axes are permitted.
The position and velocity of the joint connecting the pole to the robot are provided as observations.
The reward is 1 unless the pole is about to fall over, which terminates the episode.
A falling pole is defined by the distance along the Z axis between its lower and upper parts falling below 80\% of its length.

In~\Cref{tab:task-panoramas} we provide overview renderings of the Hurdles, Limbo, HurdlesLimbo, Stairs and Gaps environments to illustrate the positioning of objects along the course.
In these tasks, for which locomotion along the X-axis is essential, non-hierarchical baselines agents perform well with a simple shaped reward for movement along the X-axis.
For environments with randomly generated obstacles such as ours, the effectiveness of a simple ``forward'' reward has previously been observed by~\citet{heess2017emergence}.

\begin{table}[t]
\centering
\begin{tabular}{ll}
Hurdles & \includegraphics[width=0.81\textwidth,valign=m]{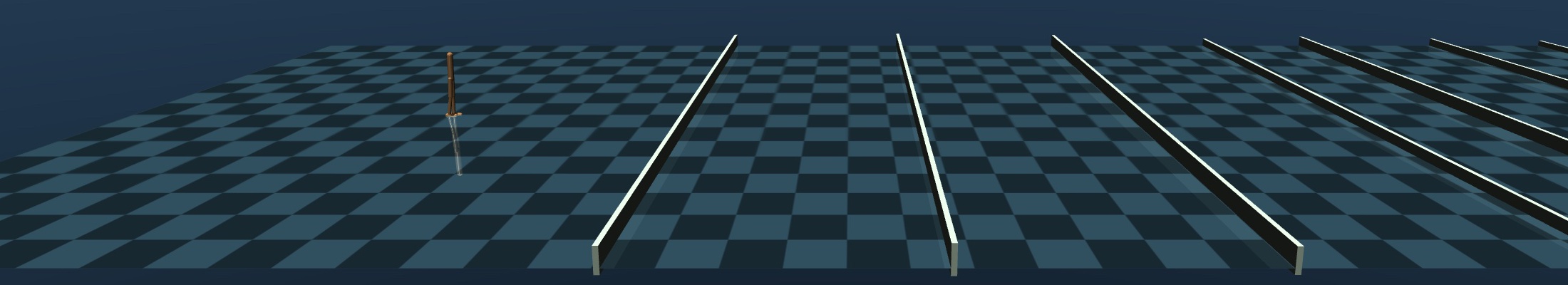} \\
Limbo & \includegraphics[width=0.81\textwidth,valign=m]{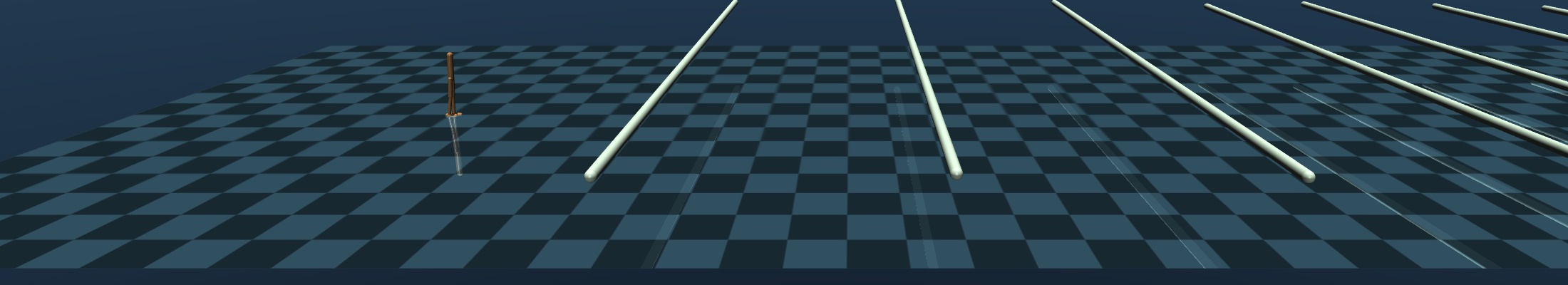} \\
HurdlesLimbo & \includegraphics[width=0.81\textwidth,valign=m]{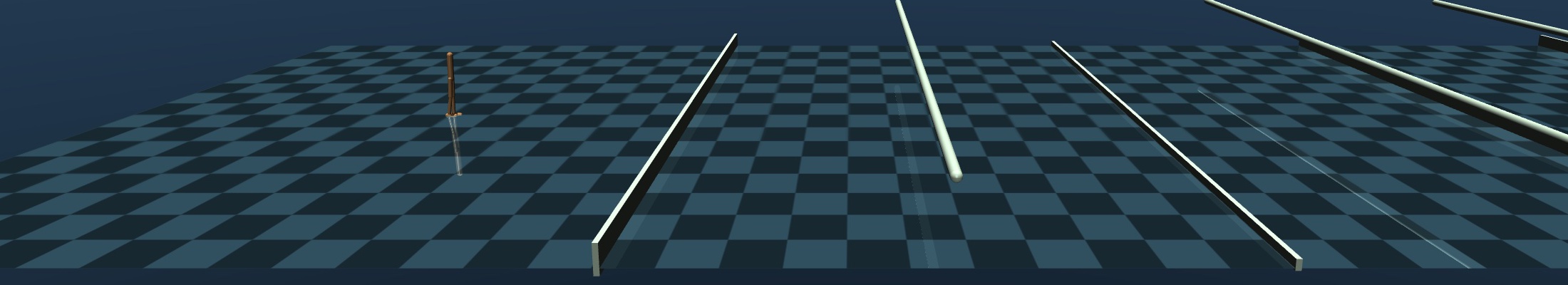} \\
Stairs & \includegraphics[width=0.81\textwidth,valign=m]{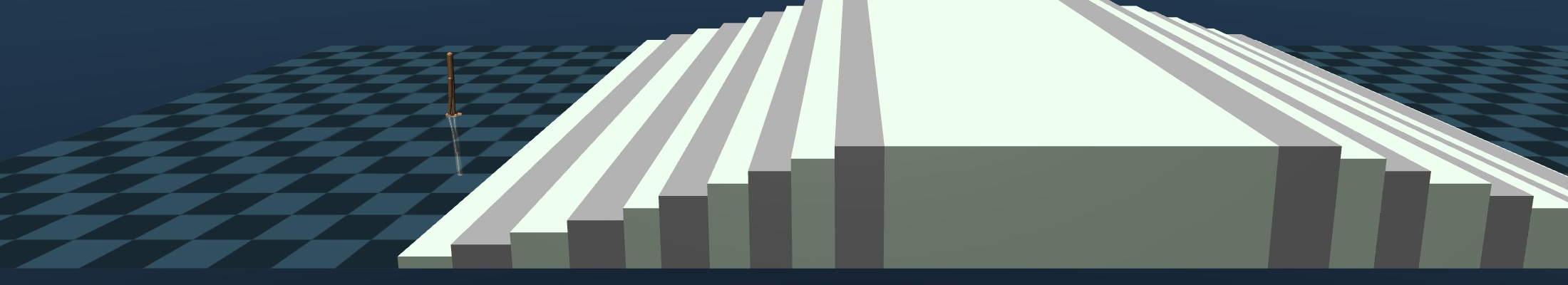} \\
Gaps & \includegraphics[width=0.81\textwidth,valign=m]{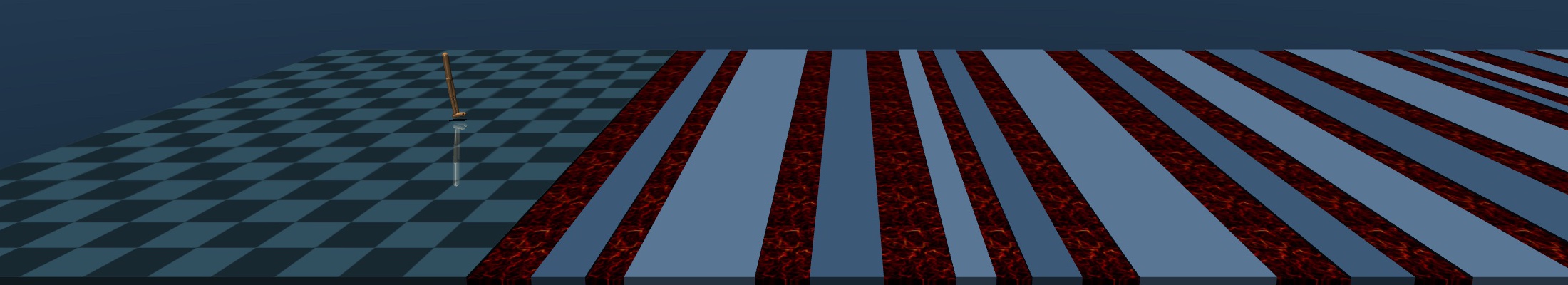} \\
\\
\end{tabular}
\caption{Rendering of 5 of our 7 benchmark tasks.
Positions of additional objects (hurdles, limbo bars, gaps, stairs) are subject to random perturbation for each episode.
Courses for Hurdles, Limbo, HurdlesLimbo and Gaps continue further.} 
\label{tab:task-panoramas}
\end{table}

%% file: app-goal-spaces.tex
\section{Robots and Goal Spaces}
\label{sec:robots-and-goal-spaces}

\subsection{Walker}
\label{sec:details-walker}

This two-dimensional, bipedal walker has been released as part of the dm\_control suite~\citep{tassa2020dm}.
We use an observation space similar to the MuJoCo tasks in Gym~\cite{brockman2016openai}, where we featurize the positions of all modeled joints, their velocities, and contact forces for all body parts clipped to $[-1,1]$.
We manually construct five primitive goal spaces that can be freely combined to obtain a total of 31 candidate goal spaces:

\begin{minipage}{\textwidth}
  \begin{minipage}[b]{0.78\textwidth}
  \adjustbox{valign=b}{
    \centering
\begin{tabular}{llll}
  \toprule
  Feature & & Range (min,max) & Direct Obs. \\
  \midrule  
  Torso X position &  & $-3,3$ & no \\
  Torso Y rotation &  & $-1.3,1.3$ & yes \\
  Torso Z position &  & $0.95,1.5$ & yes \\
  \multirow{2}{*}{Left foot, relative to torso}&
  X pos.  & $-0.72,0.99$ & no \\
  & Z pos  & $-1.3,0$ & no \\
  \multirow{2}{*}{Right foot, relative to torso}&
  X pos.  & $-0.72,0.99$ & no \\
  & Z pos.  & $-1.3,0$ & no \\
  \bottomrule
\end{tabular}
}
  \end{minipage}
  \hfill
  \begin{minipage}[t]{0.19\textwidth}
  \adjustbox{valign=b}{
    \includegraphics[width=\textwidth]{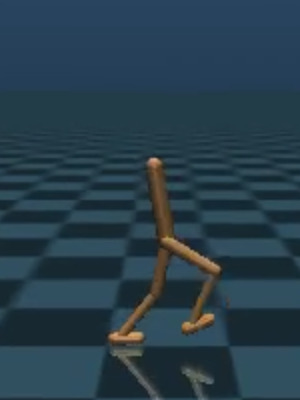}
   }
  \end{minipage}
\end{minipage}

Directly observable features are included in proprioceptive observations; non-observable ones can either be derived (i.e., the foot position can be determine via joint positions) or correspond to global observations (X position).
All position features are measured in meters while rotations are expressed in radians.
Ranges were obtained via rollouts from standard SAC policies trained with shaping rewards.

For the Walker robot, we define invalid states as the torso's Z position falling below 0.9 meters, or its Y rotation having a value outside of $[-1.4,1.4]$.
These limits are intended to prevent the robot from falling over.

\subsection{Humanoid}
\label{sec:details-humanoid}

We adopt the simplified Humanoid robot from~\citet{tassa2020dm}, which consists of 21 joints.
We use a modified observation space with the same type of state features as the Walker robot (\ref{sec:details-walker}).
Following recent work on skill learning for Humanoid robots~\citep{peng2019mcp,merel2019neural}, actions consist of target joint positions, mapped to $[-1,1]$.
We found this to produce better results compared to specifying raw torques.

For goal space featurization, closely follow our Walker setup and provide torso X and Z positions, Y rotation and relative foot positions.
Feature values for rotations around Y and Z were obtained with a twist-and-swing decomposition of the torso's global orientation matrix~\citep{dobrowolski2015swingtwist}.
State features for learning agents use a standard quaternion representation, however.

\begin{minipage}{\textwidth}
  \begin{minipage}[b]{0.75\textwidth}
  \adjustbox{valign=b}{
    \centering
\begin{tabular}{llll}
  \toprule
  Feature & & Range (min,max) & Direct Obs. \\
  \midrule  
  Torso X position &  & $-3,3$ & no \\
  Torso Y rotation &  & $-1.57,1.57$ & no \\
  Torso Z position &  & $0.95,1.5$ & yes \\
  Torso Z rotation &  & $-1.57,1.57$ & no \\
  \multirow{3}{*}{Left foot, relative to hip}&
  X pos.  & $-1,1$ & no \\
  & Y pos.  & $-1,1$ & no \\
  & Z pos.  & $-1,0.2$ & no \\
  \multirow{3}{*}{Right foot, relative to hip}&
  X pos.  & $-1,1$ & no \\
  & Y pos.  & $-1,1$ & no \\
  & Z pos.  & $-1,0.2$ & no \\
  \bottomrule
\end{tabular}
}
  \end{minipage}
  \hfill
  \begin{minipage}[t]{0.24\textwidth}
  \adjustbox{valign=b}{
    \includegraphics[width=\textwidth]{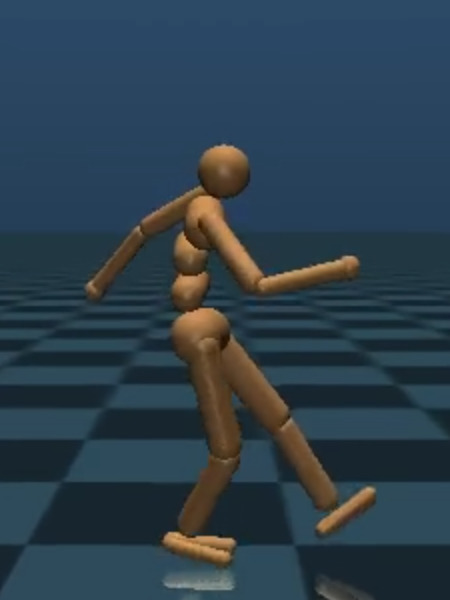}
   }
  \end{minipage}
\end{minipage}

By coupling every feature combination with the Z rotation feature, we again obtain $2^{5}-1 = 31$ possible goal spaces as for the Walker robot.

Similar to the Walker robot, we define valid states as the torso's position being at least 0.9 meters above ground.
Body rotations are not restricted.

\subsection{Goal Space Construction}
Our main considerations for formalizing goal spaces are the elimination of bias due to different feature magnitudes and a convenient notation for considering subsets of features from the goal feature space $\statesgoal$.
Starting from a set of $n \coloneqq \dim(S^g)$ features with ranges $(l^i,h^i): 1 \leq i \leq n$, we construct a goal space transformation matrix $\Psi$ along with offsets $b$:
\begin{equation*}
\Psi \coloneqq
2 I_n \begin{bmatrix} 
(h^1 - l^1)^{-1} \\
(h^2 - l^2)^{-1} \\
\vdots \\
(h^n - l^n)^{-1}
\end{bmatrix}
,\ 
b \coloneqq
-2 \begin{bmatrix} 
l^1(h^1 - l^1)^{-1} \\
l^2(h^2 - l^2)^{-1} \\
\vdots \\
l^n(h^n - l^n)^{-1}
\end{bmatrix}
- 1
\end{equation*}

A single goal space over features $F = \{i,j,k,\dots\}$ is defined as the image of an affine function $\omega^F: \statesgoal \to \goals^F$,
\begin{equation*}
\label{eq:abstraction}
\omega^F(s) \coloneqq
s
\begin{bmatrix}
\Psi_{i,*} \\
\Psi_{j,*} \\
\Psi_{k,*} \\
\vdots
\end{bmatrix}^T +
\begin{bmatrix}
b_{i} \\
b_{j} \\
b_{k} \\
\vdots
\end{bmatrix}
\end{equation*}

$\omega^F$ can be understood as simple abstraction that selects a subset $F$ of goal space features and applies a suitable normalization, mapping $[l^i,h^i]$ to $[-1,1]$ for $i \in F$.
A desirable effect of this normalization is that distance-based rewards that are commonly used to train goal-based policies will also be normalized, which facilitates optimization when sharing the parameters of policies across several goal spaces.
Additionally, the resulting high-level action space for subgoals can be conveniently defined over $[-1,1]^F$.

%% file: app-pre-training.tex
\section{Unsupervised Pre-Training Details}
\label{sec:unsup-pre-training-details}

Our unsupervised pre-training algorithm is provided in~\Cref{al:pre-training}.
We assume that the pre-training environment provides access to both proprioceptive states (the input of the skill policy) and goal state features as defined in~\Cref{sec:robots-and-goal-spaces}.
During training, goal spaces and goals are randomly selected for each episode
The low-level skill policy $\policylo$ and the corresponding Q-functions are trained with the standard Soft Actor-Critic update~\citep[Alg. 1]{haarnoja2018soft}, with representations of both goal features $F$ and goal $g$ considered a part of the state observations.

The \textsc{step\_env} function in~\Cref{al:pre-training} samples a transition from the pre-training environment, computes the reward as described in ~\Cref{sec:unsup-pre-training} and determines the values of \texttt{done} and \texttt{reset}.
\texttt{done} is set to true if one of the following conditions hold:
\begin{itemize}
  \item The current goal is reached (i.e., the L2 distance to it is smaller than $\epsilon_g$).
  \item A fixed number of steps (horizon $h$) has passed since the previous goal was sampled.
  \item A random number sampled from $[0,1)$ is smaller than the resample probability $\epsilon_r$.
\end{itemize}
\texttt{reset} is set to true if any of these conditions hold:
\begin{itemize}
  \item The state that was reached is considered invalid (\ref{sec:environment-details}). In this case, the reward is set to $-1$.
  \item A fixed number of goals (reset interval $n_r$) have been sampled without a simulation reset.
\end{itemize}

\begin{algorithm}
\caption{Unsupervised Pre-Training of Hierarchical Skills, see \Cref{fig:hsd3-system} for context}\label{al:pre-training}
\begin{algorithmic}[1]
\Require Goal spaces defined via feature sets $\mathcal{F}$ and transformations $\omega^F$
\State Pre-training environment $E$ with $\states = \statesppc \cup \statesgoal$
\State Initialize policy $\mu_\theta: \statesppc \times \mathcal{F} \times \goals \to \actions$
\State Initialize Q-function $Q_{\phi_i}: \statesppc \times \mathcal{F} \times \goals \times \actions \to \mathbb{R}$ for $i \in \{1,2\}$
\State Replay buffer $B \gets \textsc{CircularBuffer}()$
\State{reset $\gets$ true}
\For{$i\gets 1,N$}
  \If{reset}
    \State $s \sim S_0$
    \Comment{Reset simulation}
  \EndIf
  \State $F \sim \mathcal{F}, g \sim [-1,1]^F$
  \Comment{Sample new goal space and goal}
  \State{reset $\gets$ false, done $\gets$ false}
  \While{\textbf{not} (done \textbf{or} reset)}
    \State $a \sim \mu_\theta(s^p,F,(\omega^F)^{-1}(g)-s^g)$
    \State $s', r, \text{done}, \text{reset} \gets$  \textsc{step\_env}($s,a,F,g$)
    \State $B\operatorname{.append}(s^p,F,(\omega^F)^{-1}(g)-s^g,a,r,s')$
    \State $s \gets s'$
    \If{$i\%f_u == 0$}
      \For{each gradient step}
        \State $\phi, \theta \gets $ \textsc{sac\_update}($\phi,\theta,B$)
        \Comment{Perform standard SAC update}
      \EndFor
    \EndIf
  \EndWhile
  \EndFor
\State \textbf{Output:} $\theta, \phi_1, \phi_2$
\end{algorithmic}
\end{algorithm}

\begin{table}[]
  \small
  \centering
  \begin{tabular}{lll}
    \toprule
    Parameter & Value (Walker) & Value (Humanoid)\\
    \midrule
    Optimizer & Adam~\cite{kingma2015adam} & Adam \\
    Learning rate $\lambda_Q$ & $0.001$ & $0.001$ \\
    Learning rate $\lambda_\pi$ & $0.001$ & $0.001$ \\
    Learning rate $\lambda_\alpha$ & $0.001$ & $0.001$\\
    Target entropy $\targetentropy$ & $-\dim(\actions) = -6$ & $-\dim(\actions) = -21$ \\
    Initial temperature $\alpha$ & $0.1$ & $0.1$ \\ 
    Target smoothing coefficient $\tau$ & $0.005$ & $0.005$\\
    Control cost $\zeta$ & $0.01$ & $0$ \\
    Horizon $h$ & $72$ & $72$ \\
    Discount factor $\gamma$ & $1 - 1/h$ & $1 - 1/h$ \\
    Goal threshold $\epsilon_g$ & $0.1$ & $0.1$ \\
    Resample probability $\epsilon_r$ & $0$ & $0.01$ \\
    Reset interval $n_r$ & $100$ & $100$ \\
    Replay buffer size & $3\cdot10^6$ & $3\cdot10^6$ \\
    Parallel environments & $20$ & $40$ \\
    Environment steps per iteration & $1000$ & $5000$\\
    Gradient steps per iteration & $50$ & $50$ \\
    Mini-batch size & $256$ & $1024$ \\
    Warmup steps & $10^4$ & $10^4$ \\
    Total iterations & $10^4$ & $3.7 \cdot 10^4$\\\specialrule{\heavyrulewidth}{\aboverulesep}{2\belowrulesep}
  \end{tabular}
  \caption{Hyper-parameters for unsupervised pre-training.}
  \label{tab:pretraining-hps}
\end{table}

%% file: app-sac-extension.tex
\section{Hierarchical Control Details}
\label{sec:sac-extension}

\subsection{Soft-Actor Critic for HSD-3}
\label{sec:sac-extension-formulas}

Below, we provide explicit derivations for extending Soft Actor-Critic to a factorized action space $\mathcal{F} \times \mathcal{G}$ for $\policyhi$, consisting of discrete actions $F \in \mathcal{F}$ (goal space feature sets) and continuous actions $g \in \mathcal{G}^F$ (goals).
Our extension is performed according to the following desiderata:
(1) We utilize a shared critic $Q(s,F,g)$ and two policies $\policygs: \states \to \mathcal{F}$, $\policysg: \states \times \mathcal{F} \to \mathcal{G}^F$;
(2) we compute $\policygs$ for all discrete actions, which provides us with better value estimates a richer policy learning signal;
(3) separate, automatically adjusted temperature coefficients are used for the discrete-action policy ($\alpha$) and continuous-action policy ($\beta$).
As the action space of $\policysg$ is conditioned on $F$, we use separate coefficients $\beta^F$;
(4) we further normalize the entropy contributions from the different action spaces $\mathcal{G}^F$ by their dimensionality $|F|$, computing action log-probabilities as $|F|^{-1} \log \policysg(g | s, F)$.

The soft value function $V(s)$\citep[Eq. 3]{haarnoja2018soft} for acting with both policies $\policygs$ and $\policysg$ is given as follows:
\begin{align*}
V(s) &= \EE_{\substack{F \sim \policygs(\cdot | s), \\ g \sim \policysg(\cdot | s,F)}} \left[ Q(s,F,g) - \alpha \log \policygs(F | s) - \frac{\beta^F}{|F|} \policysg(g | s,F) \right] \\
\intertext{Computing the expectation over discrete actions $F$ explicitly yields}
V(s) &= \sum_{F \in \mathcal{F}} \policygs(F | s)  \left( \EE_{g \sim \policysg} \left[ Q(s,F,g) - \frac{\beta^F}{|F|} \log \policysg(g | s,F) \right] - \alpha \log \policygs(F|s) \right) \\
&= \sum_{F \in \mathcal{F}} \policygs(F | s)  \EE_{g \sim \policysg} \left[ Q(s,F,g) - \frac{\beta^F}{|F|} \log \policysg(g | s,F) \right]  + \alpha \entropy(\policygs(\cdot|s))
\end{align*}

We arrive at the formulation in~\Cref{sec:hierarchical-control} by subtracting the entropy of the uniform discrete-action policy, $\log |\mathcal{F}|$, from $\entropy(\policygs(\cdot|s))$ to ensure negative signage.
We proceed in an analogous fashion for policy~\citep[Eq. 7]{haarnoja2018soft} and temperature~\citep[Eq. 18]{haarnoja2018soft} losses.

The resulting high-level policy training algorithm is listed in~\Cref{al:high-level}.
For brevity, loss and value function definitions above use a single variable for states which indicates both proprioceptive observations $s^p$ and task-specific features $s^+$.

\begin{algorithm}
\caption{HSD-3 High-level Policy Training, see \Cref{fig:hsd3-system} for context}\label{al:sup-training}
\begin{algorithmic}[1]
\Require Low-level policy $\policylo_\theta$
\Require Goal spaces defined via feature sets $\mathcal{F}$ and transformations $\omega^F$
\Require High-level action interval $c \in \mathbb{N}^+$
\State Initialize policies $\policygs_\phi$, $\policysg_\psi$,  Q-functions $Q_{\rho_i}$ $i \in \{1,2\}$, target networks $\overline{\rho_1} \gets \rho_1, \overline{\rho_2} \gets \rho_2$
\State Replay buffer $B \gets \textsc{CircularBuffer}()$
\State $s \sim S_0,\ t \gets 0$
\For{each iteration}
  \For{each environment step}
    \If{$t \% c == 0$} 
      \Comment{Take high-level action}
      \State $F \sim \policygs_\phi([s^p;s^+])$
      \State $g \sim \policysg_\psi([s^p;s^+], F)$
      \State $g' \gets (\omega^F)^{-1}(g) - s^g$
      \Comment{Backproject goal to $\statesgoal$ and obtain delta}
    \EndIf
    \State $a \gets \EE_a \left[ \policylo_\theta(a | s^p,F,g') \right]$
    \Comment{Act with deterministic low-level policy}
    \State $s' \sim p_E(s' | s, a)$
    \Comment{Sample transition from environment}
    \State $B\operatorname{.append}(s,F,g,a,t,r(s,a),s')$
    \State $g' \gets s^g - s'^g + g'$
    \Comment{Update goal}
    \State $s \gets s'$
    \State $t \gets t+1$
    \If{end of episode}
      \State $s \sim S_0,\ t \gets 0$
    \EndIf
  \EndFor

  \For{each gradient step}
    \State Sample mini-batch from $B$, compute losses
    \Comment{See~\Cref{sec:hierarchical-control}}
    \State Update network parameters $\phi,\psi,\rho_1,\rho_2$
    \State Update temperatures $\alpha,\beta$
    \State Update target network weights $\overline{\rho}_1, \overline{\rho}_2$
      \EndFor
\EndFor
\State \textbf{Output:} $\psi, \phi, \rho_1, \rho_2$
\end{algorithmic}
\label{al:high-level}
\end{algorithm}

%% file: app-training-details.tex
\section{Training Details}
\label{sec:training-details}

\textbf{Neural Network Architecture} For all experiments, we use neural networks with 4 hidden layers, skip connections and ReLU activations~\citep{sinha2020d2rl}.
Neural networks that operate on multiple inputs (such as the skill policy, or the Q-function in HSD-3) are provided with a concatenation of all inputs.
For the Q-function in HSD-3 and HSD-Bandit, goal inputs occupy separate channels for each goal space.
The goal policy in HSD-3 $\policysg$ is modelled as a multi-head network to speed up loss computations.
Rather than receiving  an input for the selected current goal space, the output of the respective head is selected.

\textbf{Hyper-Parameters} are listed in\Cref{tab:pretraining-hps} for pre-training and~\Cref{tab:training-hps} for HSD-3 high-level policy training.
Downstream task training runs with fixed, single goal spaces (SD, SD*) use identical hyper-parameters, but do not require $\lambda_f$, $\targetentropy^f$, and $\alpha$.
Baseline runs use a learning rate of $0.003$ for neural networks and an initial temperature of $0.1$~\citep{sinha2020d2rl}.
For HSD-3 and SD, we searched for learning rates in $\{0.0001,0.0003\}$ and initial temperature values in $\{0.1,1\}$.

\textbf{Evaluation Protocol} In regular intervals (\Cref{tab:training-hps}), we perform 50 trials with a deterministic policy and measure the average return that was achieved across trials.
Since initial states and environment configurations are randomly sampled, we use the same set of 50 environment seeds for all evaluations to ensure comparability.

\subsection{Baselines}

\subsubsection{HIRO-SAC}
We implement a end-to-end HRL method similar to HIRO~\cite{nachum2018dataefficient}, but with Soft Actor-Critic as the underlying learning algorithm rather than TD3.
High-level policy actions are expressed within the goal space that we use for pre-training skill policies (\ref{sec:robots-and-goal-spaces}).
The neural network policies we use for SAC predict both mean and variances of a Gaussian distribution, and hence we perform goal relabelling by maximizing the log-probabilities of low-level actions directly instead of approximating them via squared action differences~\citep[Eq. 5]{nachum2018dataefficient}.
We observed worse performance by including DynE critic updates~\citep{whitney2020dynamicsaware}.
We therefore report results with an update schedule similar to~\citep{nachum2018dataefficient}, where high-level policies are being updated less frequently than low-level ones (depending on the high-level action frequency).

\subsubsection{DIAYN-C}
In a setup similar to~\citet{achiam2018variational}, we learn a continuous representation of DIAYN's discrete skill variable.
We input the one-hot encoded skill index to a linear layer with 7 outputs, which corresponds to $\dim(\statesgoal)$ for the walker.
Its output is subject to a hyperbolic tangent activation so that the final low-level policy skill input is bounded in $[-1,1]$.
We operate DIAYN's discriminator on our pre-defined goal spaces (\ref{sec:robots-and-goal-spaces}) and hence provide the same prior knowledge as in our methods.
After pre-training with DIAYN, we train a high-level policy as we do for SD baselines, providing its actions directly to the pre-trained policy.
We ran experiments with 256 or 1024 hidden units for the skill policy, and with 5,10,20,50 and 100 discrete skils for pre-training.
We found that best overall performance was achieved with 256 hidden units for the low-level policy and 10 different skills.

\subsubsection{Switching Ensemble}
As proposed by~\citet{nachum2019why}, this baseline consists of a small number of standard policies that gather shared experience and are randomly selected for short time-spans during rollouts.
In our version, we use SAC as the underlying learning algorithm, in the same configuration as for the SAC baseline.
We use an ensemble of 5 policies, and perform switches with the same frequency that high-level actions are taken at for the other hierarchical methods.
For evaluations, we act with a single policy throughout the entire episode.

\subsubsection{HIDIO}
We use the official implementation from Github\footnote{\url{https://github.com/jesbu1/hidio/tree/245d758}}.
We found it non-trivial to present the low-level policy discriminator with observations in our standard goal space.
Hence, in contrast to HIDIO-SAC and DIAYN-C, the HIDIO baseline discriminator operates on the full state space.
In accordance with the other hierarchical methods considered, the steps per option were set to 1 on PoleBalance and 5 otherwise, and the discount factor was set to 0.99.
Likewise, we use 256 hidden units for the low-level policy's neural network layers.
We performed a hyper-parameter sweep on the Hurdles task, similar to the one performed in the original paper~\citep{zhang2020hierarchical} (\Cref{tab:hidio-sweep}).

\begin{table}[h]
  \small
  \centering
  \begin{tabular}{ll}
    \toprule
    Parameter & Value \\
    \midrule
    Discriminator input & state, action, state\_action, \textbf{state\_difference} \\
    Latent option vector dimension (D) & \textbf{8}, 12  \\
    Rollout length & 25, 50, \textbf{100}  \\
    Replay buffer length & 50000, \textbf{200000}\\\specialrule{\heavyrulewidth}{\aboverulesep}{2\belowrulesep}
  \end{tabular}
  \caption{Hyper-parameters considered for HIDIO, with best ones emphasized.}
  \label{tab:hidio-sweep}
 \end{table}

\begin{table}[]
  \small
  \centering
  \begin{tabular}{lll}
    \toprule
    Parameter & Value (Walker) & Value (Humanoid)\\
    \midrule
    Optimizer & Adam~\cite{kingma2015adam} & Adam \\
    Learning rate $\lambda_Q$ & $0.001$ & $0.001$ \\
    Learning rate $\lambda_f$ & $0.003$ & $0.001$ \\
    Learning rate $\lambda_g$ & $0.003$ & $0.003$  \\
    Learning rate $\lambda_\alpha$ & $0.001$ & $0.001$\\
    Learning rate $\lambda_\beta$ & $0.001$ & $0.001$ \\
    Target entropy $\targetentropy^f$ & $0.5 \log |\mathcal{F}|$ & $0.5 \log |\mathcal{F}|$ \\
    Target entropy $\targetentropy^g$ & $-1$ & $-1$ \\
    Initial temperatures $\alpha, \beta^F$ & $1$ & $1$ \\ 
    Target smoothing coefficient $\tau$ & $0.005$ & $0.005$\\
    Discount factor $\gamma$ & $0.99$ & $0.99$ \\
    Replay buffer size & $10^6$ & $2\cdot10^6$ \\
    Parallel environments & $1$ & $5$ \\
    Environment steps per iteration & $50$ & $500$\\
    Gradient steps per iteration & $50$ & $50$ \\
    Mini-batch size & $256$ & $512$ \\
    Warmup steps & $10^3$ & $10^4$ \\
    Evaluation interval (iterations) & $1000$ & $400$\\\specialrule{\heavyrulewidth}{\aboverulesep}{2\belowrulesep}
  \end{tabular}
  \caption{Hyper-parameters for high-level policy training with HSD-3.}
  \label{tab:training-hps}
\end{table}

\newpage

\section{Extended Results}
\label{sec:extended-results}

Below, we provide full learning curves for the results presented in~\Cref{sec:experiments} and additional experiments.

\subsection{Walker Learning Curves}
\label{sec:learning-curves-walker}

Learning curves for baselines, HSD-3, HSD-Bandit and SD are provided in~\Cref{tab:walker-perf-full}.
In addition to the discussion in the experimental section, these plots emphasize that exploration is challenging in all environments apart from PoleBalance.
For SAC, SE and HIRO-SAC in particular, positive returns are obtained with few seeds only (gray lines).

\begin{table}[ht]
  \centering
  \small
  \resizebox{\textwidth}{!}{
  \begin{tabular}{lrrrrrrr}
    \toprule
    Method & \hspace{1em}Hurdles\hspace{1em} & Limbo & HurdlesLimbo & Stairs & \hspace{0.7em}GoalWall & \hspace{1.2em}Gaps\hspace{0.2em} & PoleBalance \\
    \midrule
    HSD-3 & \multicolumn{7}{l}{\includegraphics[width=0.81\textwidth,valign=m]{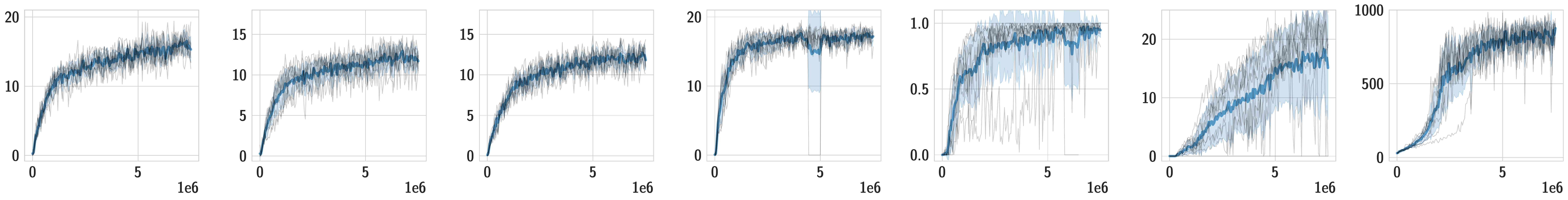}} \\
    HSD-Bandit & \multicolumn{7}{l}{\includegraphics[width=0.81\textwidth,valign=m]{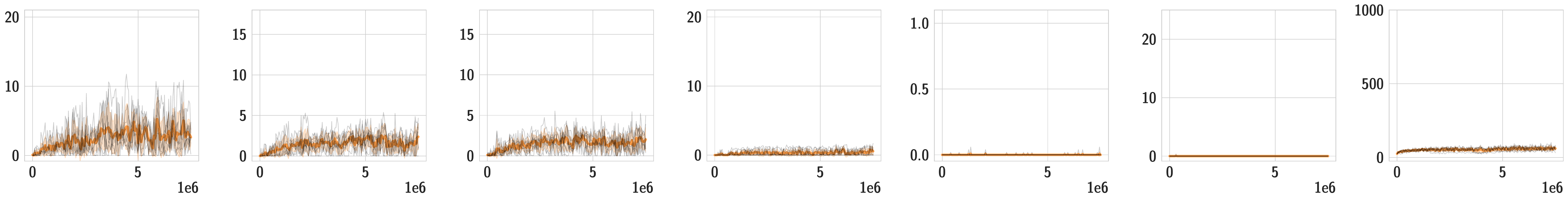}} \\
    SD & \multicolumn{7}{l}{\includegraphics[width=0.81\textwidth,valign=m]{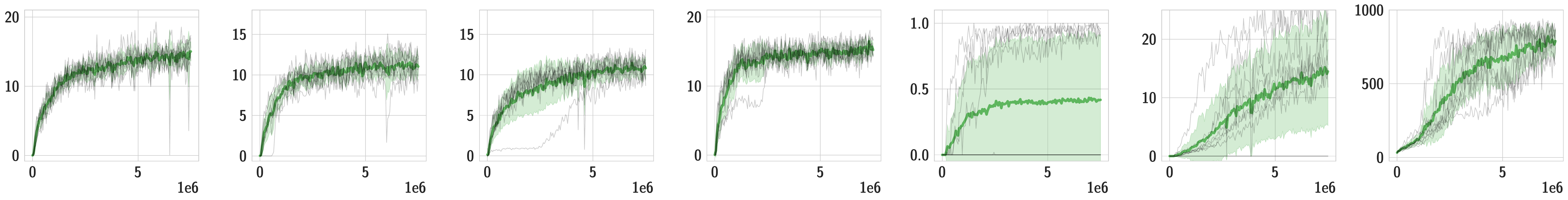}} \\
    SD* & \multicolumn{7}{l}{\includegraphics[width=0.81\textwidth,valign=m]{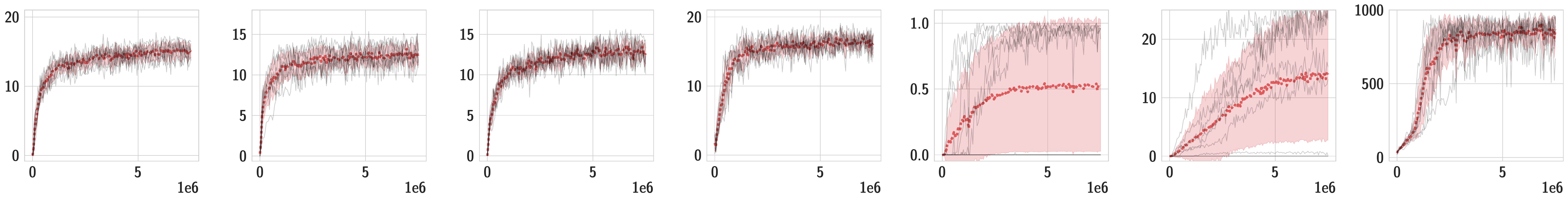}} \\
    \midrule
    SAC & \multicolumn{7}{l}{\includegraphics[width=0.81\textwidth,valign=m]{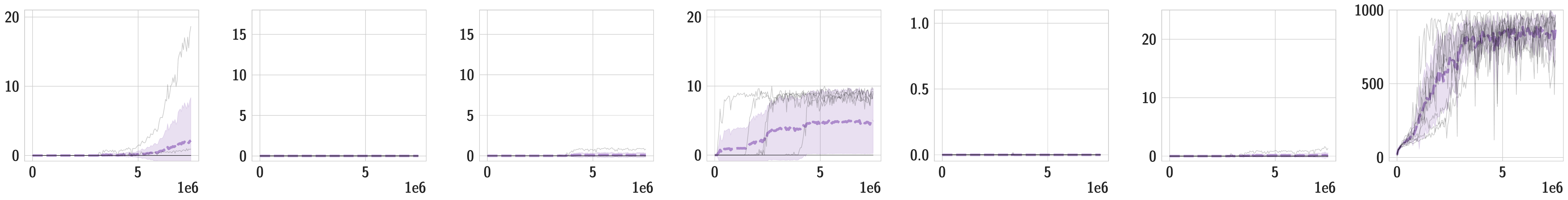}} \\
    Sw-Ensem. & \multicolumn{7}{l}{\includegraphics[width=0.81\textwidth,valign=m]{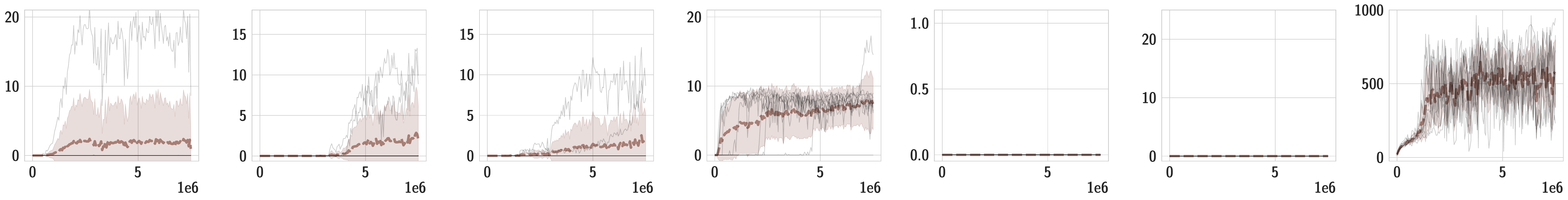}} \\
    HIRO-SAC & \multicolumn{7}{l}{\includegraphics[width=0.81\textwidth,valign=m]{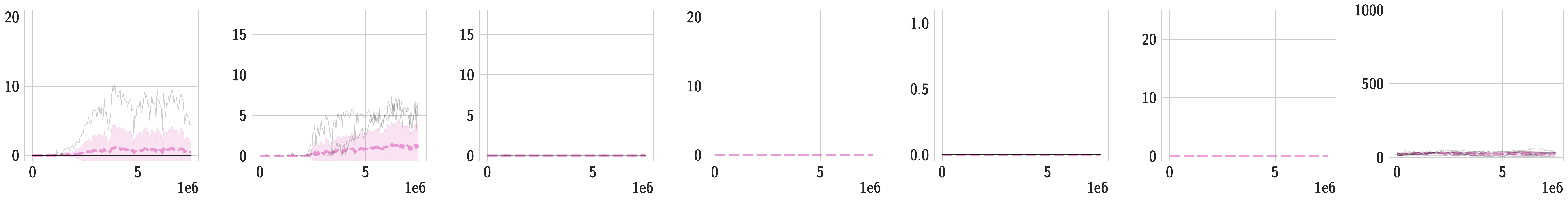}} \\
    HIDIO & \multicolumn{7}{l}{\includegraphics[width=0.81\textwidth,valign=m]{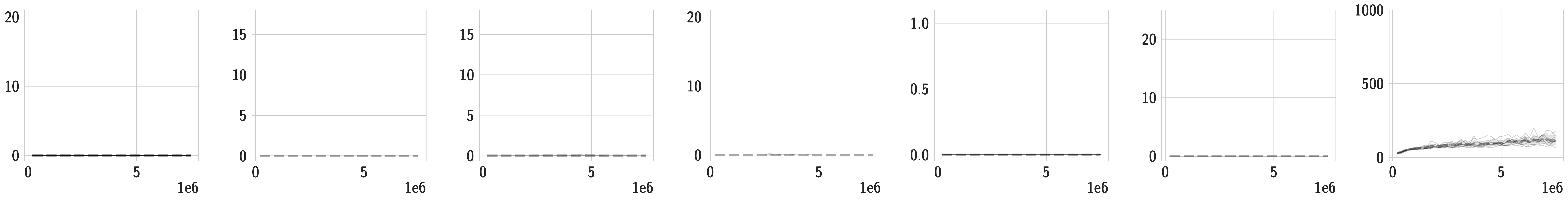}} \\
    DIAYN-C & \multicolumn{7}{l}{\includegraphics[width=0.81\textwidth,valign=m]{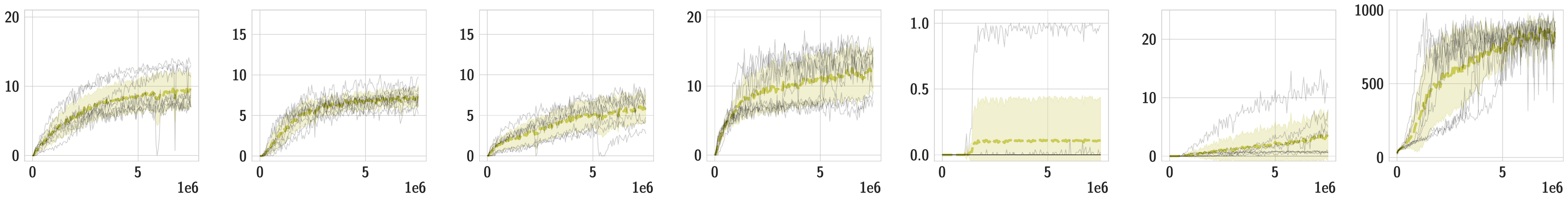}}\\\specialrule{\heavyrulewidth}{\aboverulesep}{2\belowrulesep}
  \end{tabular}
  }
  \caption{Full learning curves for results reported in~\Cref{tab:walker-perf} (Walker). We show mean returns achieved (Y axis) after interacting with an environment for a given number of steps (X axis). Shaded areas mark standard deviations computed over 9 seeds for each run.}
  \label{tab:walker-perf-full}
\end{table}

\subsection{Ablation: Multi-Task Pre-Training}
\label{sec:ablation-pre-training}

\begin{figure}
  \centering
  \includegraphics[width=\textwidth]{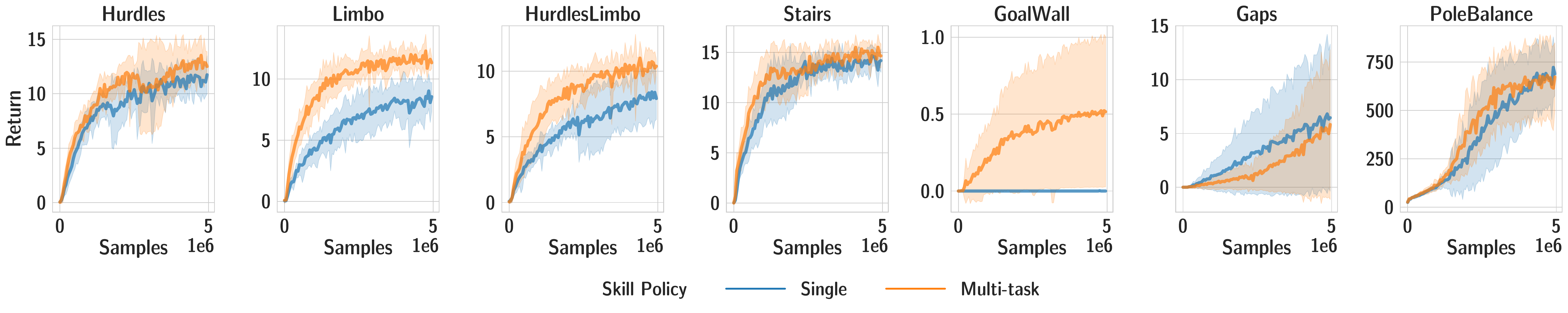}
  \caption{Returns achieved after 5M samples with the SD baseline and the Walker robot, using low-level policies that were trained on a single goal space or with our proposed multi-task pre-training scheme.}
  \label{fig:walker-ablation-multitask}
\end{figure}

We perform an ablation study on the impact of our multi-task pre-training algorithm used to obtain a hierarchy of skills (\ref{sec:unsup-pre-training-details}).
We train SD high-level policies, i.e., with a goal space consisting of all considered features, with a skill policy that was trained to reach goals defined in this goal space only.
This is in contrast to the pre-trained models used in the other experiments throughout the paper, which are trained to achieve goals withih a hierarchy of goal spaces.
Networks used to train the single goal-space skill policy consist of 256 units per hidden layer, while skill policies shared among multiple goal spaces use 1024 units (increasing the number of parameters for the single skill policy resulted in worse performance).
The results in~\Cref{fig:walker-ablation-multitask} show that shared policies obtained with multi-task pre-training yield higher returns in most benchmark environments.
For GoalWall, the usage of a single-skill policy prevented learning progress altogether.
These findings indicate the effectiveness of the multi-trask pre-training stage: it not only produces a compact representation of various skills, making downstream usage more practical, but also results in improved policies for individual goal spaces.

\subsection{Analysis: Exploration Behavior}
\label{sec:exploration-behavior}

\begin{figure}[t]
  \centering
  \includegraphics[width=\textwidth]{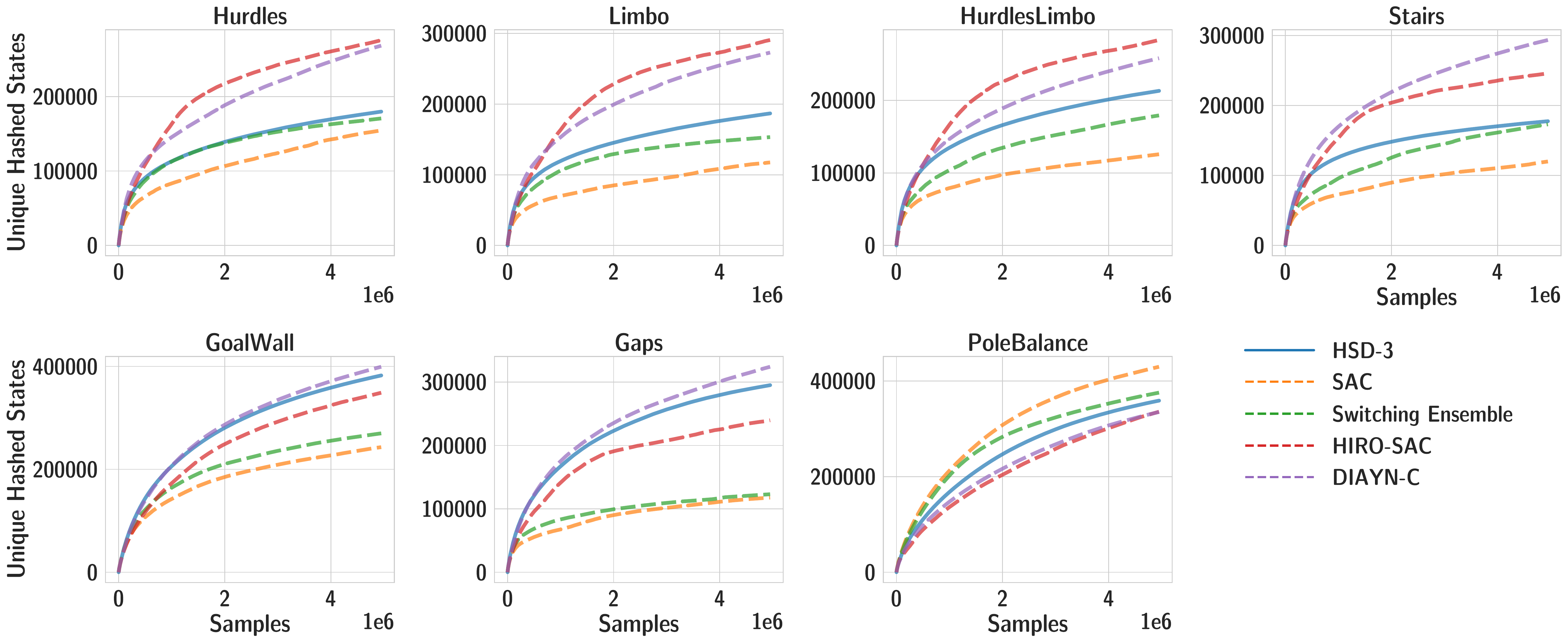}
  \caption{Number of unique hashed states encountered over the course of training (Walker). Mean over 9 seeds per task and method.}
  \label{fig:exploration-states}
\end{figure}

For gaining additional insight into how HSD-3 impacts exploration behavior, we analyze state visitation counts over the course of training.
Due to the continuous state spaces of our environments, we estimate the number of unique states visited over the course of training with SimHash, a hashing method originally proposed for computing exploration bonuses~\citep{charikar2002similarity,tang2017exploration}.
We hash the full observation, which can include other, randomly placed objects (such as Hurdles or a ball).
We compare the amount of unique hashed states for HSD-3 and selected baselines with the Walker robot in~\Cref{fig:exploration-states}.
Generally, hierarchical methods encounter more states during training compared to SAC, even if this does not necessarily translate to a higher return (cf. \Cref{fig:walker-perf}).
For HSD-3, after an initial phase of fast learning in Hurdles, Limbo, HurdlesLimbo and Stairs, the performance in terms of return converges, which is reflected in a decrease of new states being visited.

\subsection{Analysis: Pre-Training Performance}
\label{sec:walker-pretraining-curves}

\begin{figure}[h]
  \centering
  \includegraphics[width=\textwidth]{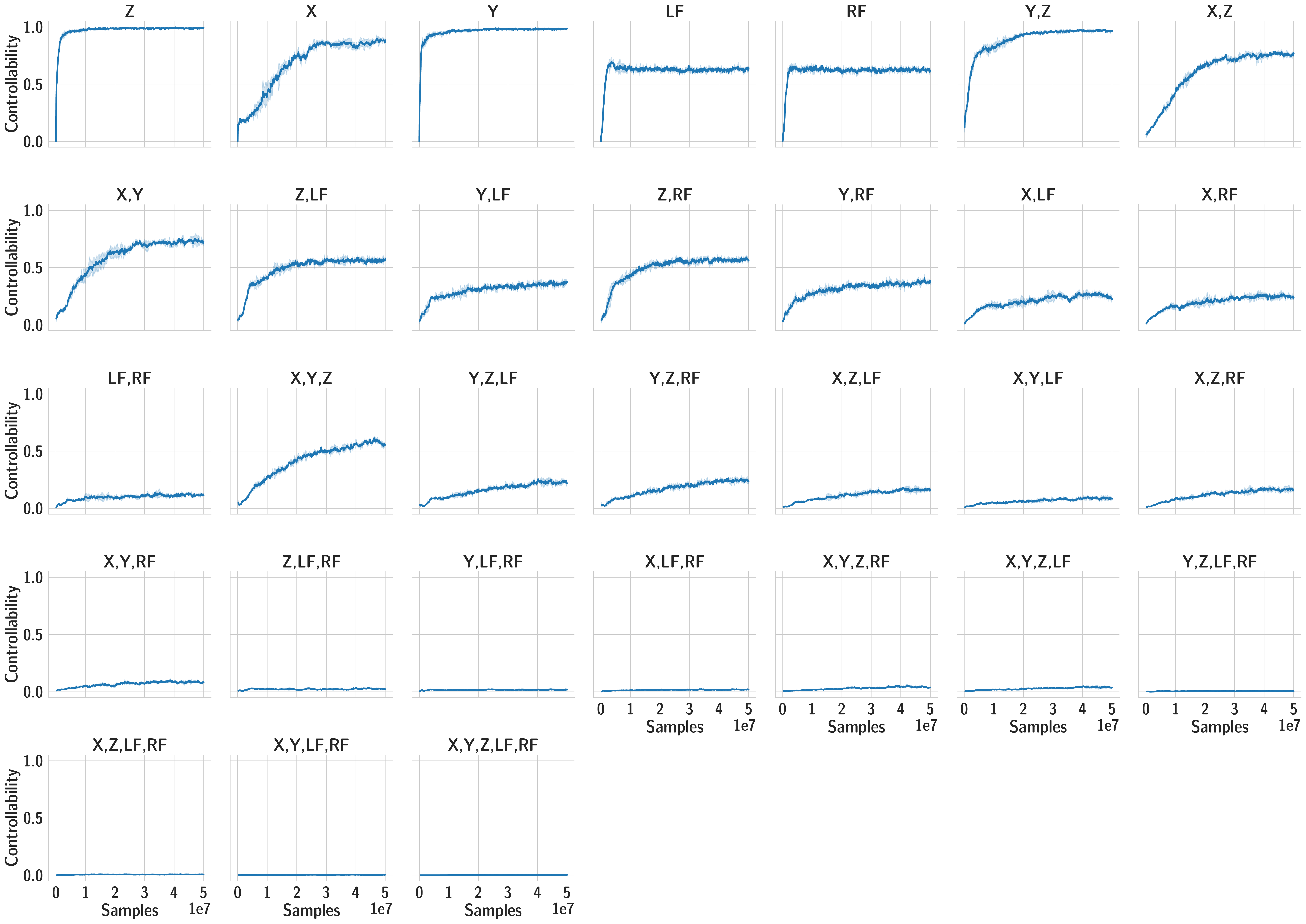}
  \caption{Pre-training performance over the different feature sets considered (Walker robot). Controllability (Y axis) is estimated with a dedicated model. Mean and standard deviation over 3 runs.}
  \label{fig:walker-pretraining-ctrlb}
\end{figure}

In~\Cref{fig:walker-pretraining-ctrlb}, we plot performance during unsupervised pre-training.
We train a set goal-reaching policies for different feature combinations, modeled with a single neural network.
We found that in our pre-training setup, the training reward alone does not adequately capture the fraction of goals that can be reached reliably.
As the number of  feature subsets increases, dedicated evaluation runs needlessly prolong the wall-clock training time.
As an alternative measure of performance, we train an additional Q-function on a 0/1 reward (1 if a goal was reached) and query it periodically with initial states from the replay buffer and randomly sampled goals.
The resulting metric reflects controllability, i.e., the probability with which a randomly sampled goal can be reached.

\Cref{fig:walker-pretraining-ctrlb} shows that, with an increasing number of features, goals become harder to reach.
While it is unsurprisingly harder to achieve goals in many dimensions, another effect is that we do not account for fundamentally unreachable states in our definition of the goal space $S^g$.
For example, the reachable states for the feet (LF, RF), which are two-dimensional features (X and Z position) roughly describe a half moon while we sample goals from a hypercube.
This effect is multiplied when feature sets are combined.

\subsection{Humanoid Learning Curves}
\label{sec:learning-curves-humanoid}

\begin{figure}
  \centering
  \includegraphics[width=\textwidth]{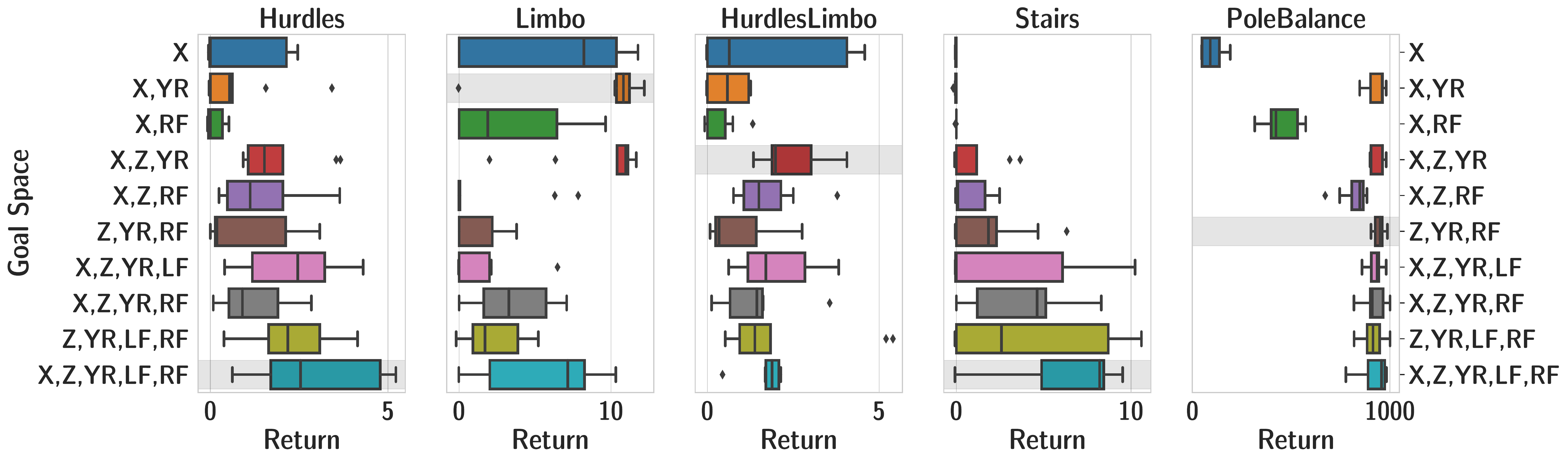}
  \caption{Returns achieved after 50M samples on the benchmark tasks with the Humanoid robot with fixed low-level policy goal spaces (individual runs and quartiles). Each row corresponds to a set of features for the respective goal space. All goal spaces also include the Z rotation of the torso as a feature.}
  \label{fig:humanoid-single}
\end{figure}

In~\Cref{fig:humanoid-single}, we plot performance for high-level policy training with fixed, individual goal spaces on the Humanoid robot.
Similar to the results for the Walker robot (\Cref{fig:walker-single}), the best goal space features differ significantly across tasks.

Learning curves with the Humanoid robot are presented in~\Cref{tab:humanoid-perf}.
The non-hierarchical SAC baseline achieves positive returns in the Stairs environment only.
The two successful runs manage to climb up the initial flight of stairs but fall over at the top, thus not making it through the entire course.
The mediocre median performance of HSD-3 on Stairs is attributed to one of the three seeds failing to learn; the other two are on par with the SD baseline.
Notably, HSD-3 outperforms the best single goal space on the Limbo and HurdlesLimbo tasks.

\begin{table}
  \centering
  \small
  \resizebox{\textwidth}{!}{
  \begin{tabular}{lrrrrc}
    \toprule
    Method & \hspace{2em}Hurdles\hspace{1.5em} & \hspace{1.5em}Limbo\hspace{1em} & \hspace{1em}HurdlesLimbo\hspace{0.7em} & \hspace{1.5em}Stairs\hspace{1.7em} & PoleBalance \\
    \midrule
    SAC & \multicolumn{5}{l}{\includegraphics[width=0.81\textwidth,valign=m]{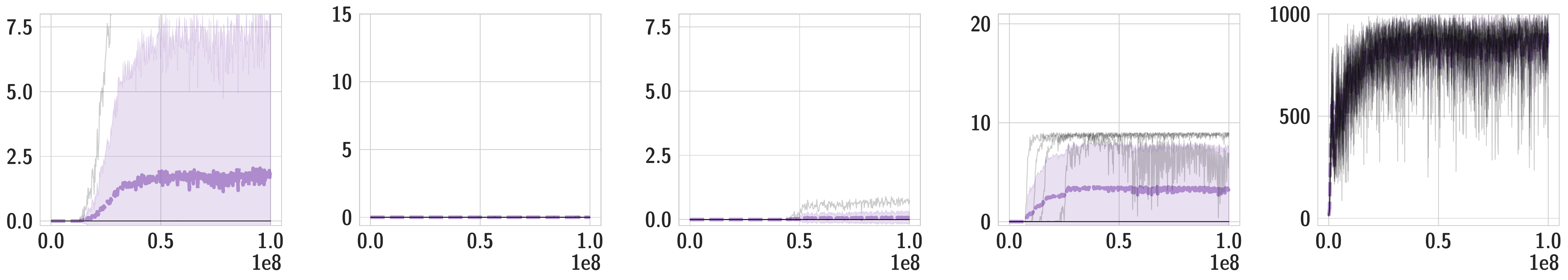}} \\
    \midrule
    HSD-3 & \multicolumn{5}{l}{\includegraphics[width=0.81\textwidth,valign=m]{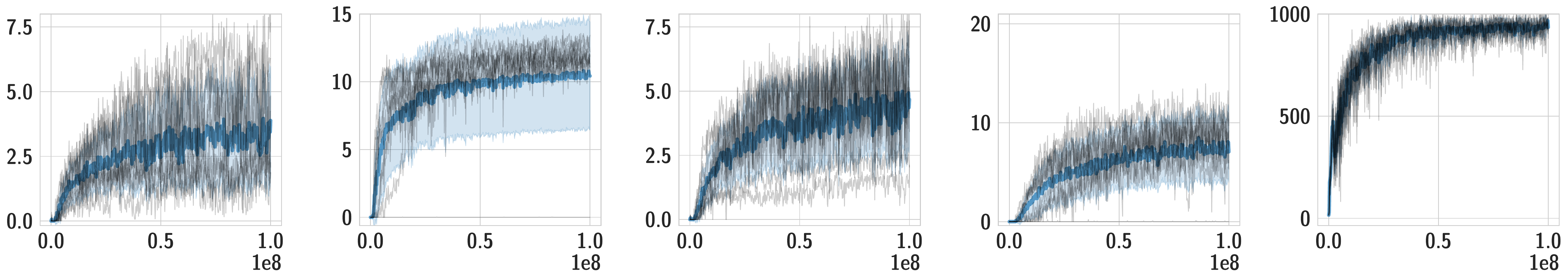}} \\
    SD & \multicolumn{5}{l}{\includegraphics[width=0.81\textwidth,valign=m]{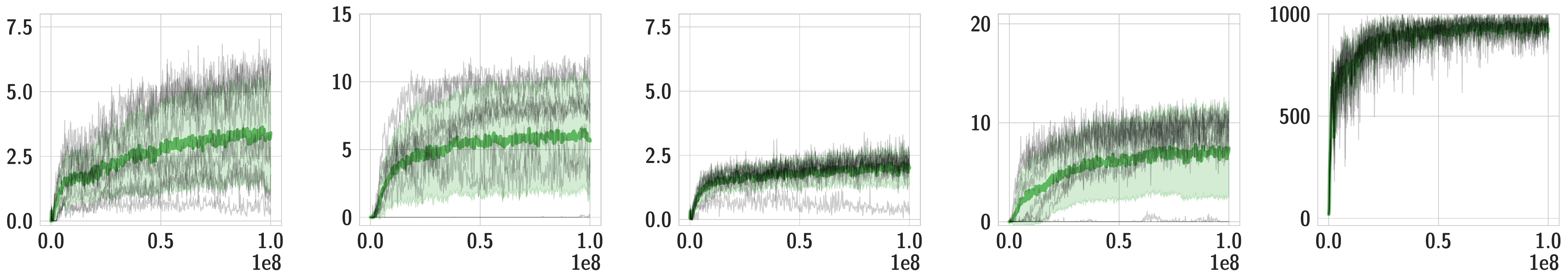}} \\
    SD* & \multicolumn{5}{l}{\includegraphics[width=0.81\textwidth,valign=m]{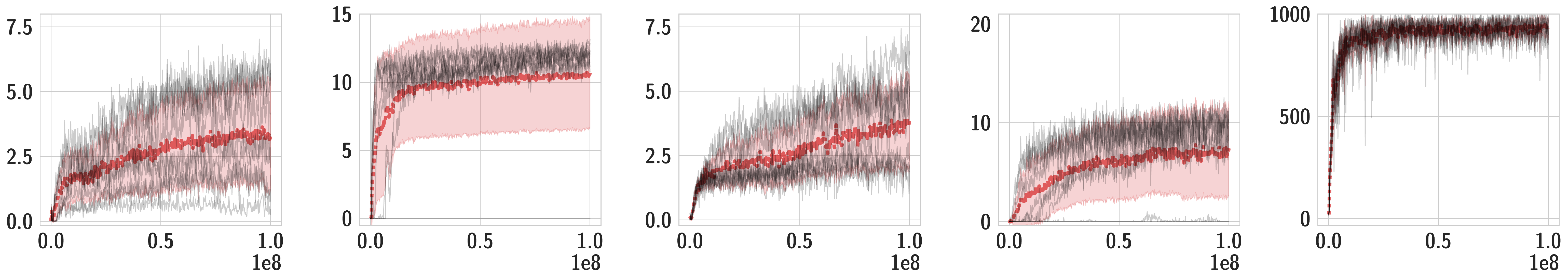}}\\\specialrule{\heavyrulewidth}{\aboverulesep}{2\belowrulesep}
  \end{tabular}
  }
  \caption{Full learning curves for results reported in~\Cref{fig:humanoid-results} (Humanoid). We show mean returns achieved (Y axis) after interacting with an environment for a given number of steps (X axis). Shaded areas mark standard deviations computed over 9 seeds for each run.}
  \label{tab:humanoid-perf}
\end{table}

%% file: app-hiro-rewards.tex
\newpage
\section{Implicit Priors in HIRO}
\label{sec:hiro-reward-normalization}

\begin{figure}[h]
  \centering
  \includegraphics[width=\textwidth]{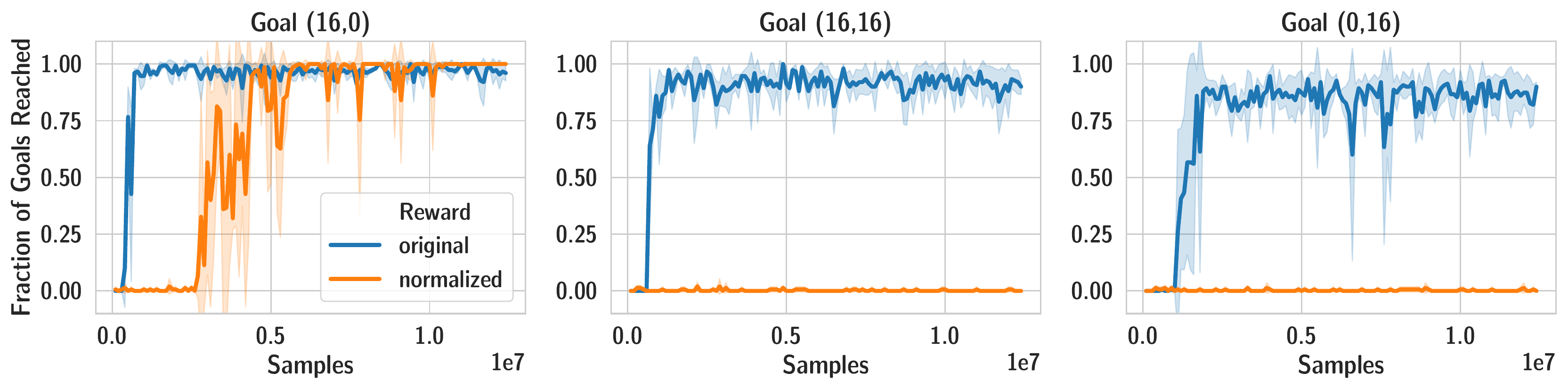}
  \caption{Fraction of goals reached in the AntMaze task with HIRO during evaluations (mean and standard deviation over 3 seeds).
  With normalized rewards, locomotion on the X-Y plane is no longer subject to higher rewards, and only the nearby goal at position (16,0) can be reached.}
  \label{fig:hiro-reward-normalization}
\end{figure}

To highlight the prevalence of implicit priors towards navigation tasks in existing methods, we perform an ablation study on HIRO, a prominent end-to-end HRL method~\cite{nachum2018dataefficient,nachum2019nearoptimal}.
HIRO defines manual subgoal ranges to steer its low-level policy, where ranges for X and Y locations are of higher magnitude compared to joint angles.
In~\Cref{fig:hiro-reward-normalization}, we analyze the impact of normalizing the reward on the AntMaze task.
We compute the intrinsic reward for the low-level policy~\citep[Eq.3]{nachum2018dataefficient} as
\begin{equation*}
r(s_t,g_t,a_t,s_{t+1}) = -\left\lVert \frac{s_t - g_t + s_{t+1}}{R} \right\rVert_2 ,
\end{equation*}
with $R = [10,10,0.5,1,1,1,1,0.5,0.3,0.5,0.3,0.5,0.3,0.5,0.3]$ corresponding to the high-level policy's action scaling~\citep[C.1]{nachum2018dataefficient}.
The normalization causes all dimensions of the goal space to contribute equally to the low-level policy's learning signal, and effectively removes any implicit prior towards center-of-mass translation.
As a result, among the three goals used for evaluation, only the nearby one at position (16,0) can be successfully reached without reward normalization.

\begin{figure}[h]
  \centering
  \includegraphics[width=\textwidth]{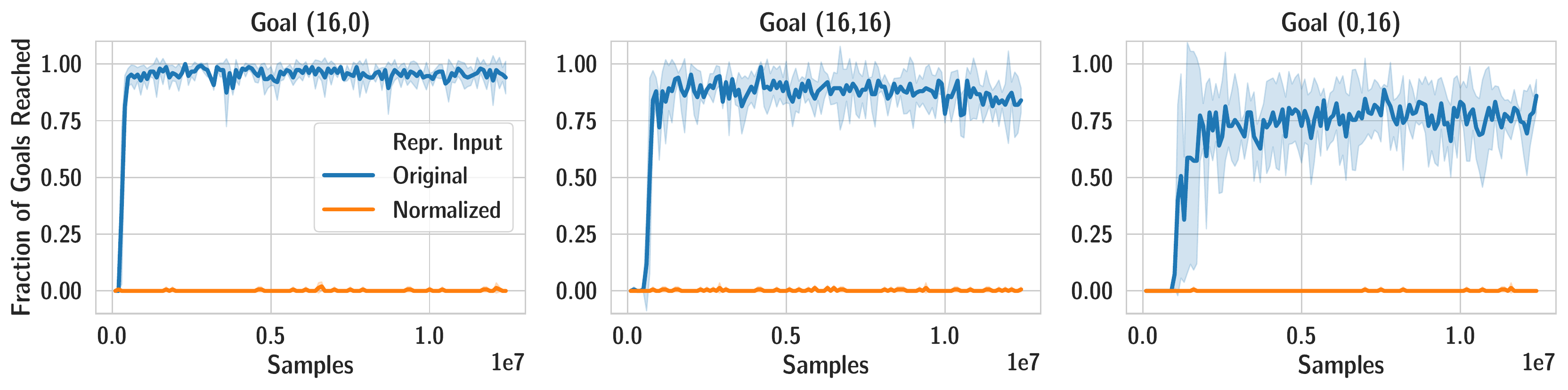}
  \caption{Fraction of goals reached in the AntMaze task with HIRO and representation learning during evaluations (mean and standard deviation over 3 seeds).
  With normalized state representation inputs, no learning progress is made.}
  \label{fig:hiro-repr-reward-normalization}
\end{figure}

\citet{nachum2019nearoptimal} propose to automatically learn a state representation used as the goal space for high-level actions.
We investigate the dependence of inductive bias towards X-Y translation by modifying their setup as follows.
First, we limit the input to the state representation function $f$~\citep[Sec. 2]{nachum2019nearoptimal} to the same 15 features that serve as a goal space in~\citep{nachum2018dataefficient}.
We then normalize the input of $f$ as above via division by $R$, and modify the high-level action scaling accordingly (actions in $[-1,1]^2$, and a Gaussian with standard derivation $0.5$ for exploration).
As a result, the agent is unable to reach any of the tested goals (\Cref{fig:hiro-repr-reward-normalization}).
The limitation of input features alone does not affect performance (blue curve).

